\documentclass[review]{elsarticle}
\usepackage[letterpaper,top=2.5cm,bottom=2.5cm,left=2.6cm,right=2.6cm,marginparwidth=1.75cm]{geometry}
\usepackage{lineno,amsmath,amsfonts,amssymb,amsthm,mathrsfs}
\usepackage[colorlinks=true,linkcolor=blue]{hyperref}
\usepackage[linesnumbered,ruled,vlined,onelanguage]{algorithm2e}
\usepackage{booktabs}
\usepackage{tikz}
\usepackage{longtable}
\usepackage[skip=0pt]{caption,subcaption}
\usepackage{float}
\usepackage{setspace}
\newtheorem{theorem}{Theorem}
\biboptions{authoryear}
\begin{document}
\title{Modeling Stochastic Microscopic Traffic Behaviors: a Physics Regularized Gaussian Process Approach}
\author{Yun Yuan, Ph.D.}
\address{Postdoctoral Research Associate, Department of Civil and Environmental Engineering, University of Utah 110 Central Campus Drive, Room 1650, Salt Lake City, UT 84112, USA
Tel: +1 414 458 3343, Email: yun.yuan@utah.edu}

\author{Qinzheng Wang}
\address{Research Assistant, Department of Civil and Environmental Engineering, University of Utah 110 Central Campus Drive, Room 1650, Salt Lake City, UT 84112, USA
Tel: +1 385 299 5786, Email: qinzheng.wang@utah.edu}

\author{Xianfeng Terry Yang*, Ph.D.}
\address{Assistant Professor, Department of Civil and Environmental Engineering, University of Utah 110 Central Campus Drive, Room MCE2000, Salt Lake City, UT 84112, USA
Tel: +1 801 585 1290, Email: x.yang@utah.edu}

\begin{abstract}
Modeling stochastic traffic behaviors at the microscopic level, such as car-following and lane-changing, is a crucial task to understand the interactions between individual vehicles in traffic streams. Leveraging a recently developed theory named physics regularized Gaussian process (PRGP), this study presents a stochastic microscopic traffic model that can capture the randomness and measure errors in the real world.  Physical knowledge from classical car-following models is converted as physics regularizers, in the form of shadow Gaussian process (GP), of a multivariate PRGP for improving the modeling accuracy. More specifically, a Bayesian inference algorithm is developed to estimate the mean and kernel of GPs, and an enhanced latent force model is formulated to encode physical knowledge into stochastic processes. Also, based on the posterior regularization inference framework, an efficient stochastic optimization algorithm is developed to maximize the evidence lower-bound of the system likelihood. To evaluate the performance of the proposed models, this study conducts empirical studies on real-world vehicle trajectories from the NGSIM dataset. Since one unique feature of the proposed framework is the capability of capturing both car-following and lane-changing behaviors with one single model, numerical tests are carried out with two separated datasets, one contains lane-changing maneuvers and the other doesn't. The results show the proposed method outperforms the previous influential methods in estimation precision.  
\end{abstract}
\begin{keyword}
microscopic traffic flow model\sep physics regularized Gaussian process\sep car-following model\sep posterior regularization inference
\end{keyword}

\maketitle

\section{Introduction}
\subsection{Background}
With the increase of travel demand, traffic congestion has become a critical issue which not only causes transportation problems (e.g., longer travel time) but also brings increased environmental pollution (e.g., vehicle emissions). Also, nonhomogeneous driver behaviors in the traffic stream play as a key role in contributing to traffic jams. Hence, modeling the stochastic nature of traffic behaviors, at the microscopic level, is essential for traffic engineers to evaluate various traffic operational functions. In the literature, one of the core models, the car-following model,
aims to describe the interactions of leading and following vehicles in the traffic. Over the past decades, numerous car-following models have been developed by transportation researchers, including both mathematical models \citep{moridpour2010lane, mardiati2014review, aghabayk2015state, mahapatra2018parametric} and data-driven models \citep{ossen2005car,rahman2013review}. Mathematical car-following models often rely on the physical knowledge of vehicle motions and describe a driver's behavior according to the speeds and locations of other nearby vehicles. In contrast, data-driven models precisely mimic the vehicle behaviors based on the analysis of a mass of empirical data (e.g., vehicle trajectories) without manual interventions. 

By comparing the performances of both types of models, mathematical car-following models usually lack the flexibility to simulate the nonhomogeneous nature of driver behaviors due to the closed-form formulations and some strong assumptions. Their model parameters also require tremendous efforts to calibrate. Data-driven car-following models directly study the correlations of interacted vehicles by analyzing empirical data. Despite some existing studies prove the effectiveness, their performance will be inevitably affected by the amount and quality of available data. Moreover, one main reason that prevents wide implementations of data-driven models is that users cannot interpret the results when encountering problems, as the models are treated as "black boxes". Meanwhile, other core microscopic traffic behavioral models, such as lane-changing models, are often formulated with probabilistic and discretized forms\citep{kesting2007general,laval2008microscopic,gipps1986model,toledo2005lane}. Their model parameters are even more difficult to calibrate, compared with car-following models, since the number of lane-changing maneuvers is relatively small in a training dataset that describes vehicle trajectories.

Hence, to fill the current research gap, this study aims to develop a novel modeling framework, named physics regularized Gaussian process (PRGP), to fuse the mathematical car-following (i.e., physics) models with a multivariate Gaussian process (GP). Then the difficulty of modeling lane-changing behaviors can be automatically handled by the data-driven (i.e., GP) part. The proposed PRGP model incorporates both information from empirical data and the a priori physical knowledge from car-following models to estimate vehicle behaviors, especially under critical situations. More specifically, the PRGP approach can 1) offer a new way to interpret model results that goes beyond the capability of data-driven models; and 2) address the limitations of mathematical methods in modeling complex and uncertain traffic systems. Besides, in the model training process, control parameters in the integrated physics (i.e., car-following) model will be automatically calibrated and the weighting factors that determine the portions of contributions from the physics and GP models will be updated over time to ensure the high performance of prediction accuracy. 

\subsection{Modeling framework}
Among existing data-driven methods, the Gaussian process (GP) \citep{williams2006gaussian} is a powerful nonparametric function estimator and has numerous successful applications.
It can capture the relationship between stochastic variables without requiring strong assumptions (such as memorylessness). 
In traffic modeling, GP-based methods have been applied in traffic speed imputation \citep{rodrigues2018heteroscedastic,rodrigues2018multi}, public transport flows \citep{neumann2009stacked}, traffic volume estimation and prediction \citep{xie2010gaussian}, travel time prediction \citep{ide2009travel}, driver velocity profiles \citep{armand2013modelling} and traffic congestion \citep{liu2013adaptive}.

However, similar to other data-driven approaches, GPs can perform poorly when the training data are scarce and insufficient to reflect the complexity of the system or testing inputs contains many noises. As the full-scale vehicle trajectories are often collected by video cameras and obtained by computer vision algorithms (e.g., NGSIM), such data issues (i.e., insufficiency and noisy) commonly exist. Therefore, scholars start to incorporate physical knowledge, from mathematical formulations, into the learning process of GPs to improve their performance.
Using GP to represent physical knowledge has the two major difficulties: (a) differential equations of physics models are hard to be represented as a probabilistic term, such as priors and likelihoods; and (b) in practice, physical knowledge is usually incomplete but the differential equations can include latent functions and parameters (e.g. unobserved noise, inflows, outflows), making their presentations and joint estimation with GPs even more challenging.
To represent differential equations in GPs \citep{alvarez2009latent, alvarez2013linear}, the proposed Latent Force Models (LFM) for training and prediction of GP are based on the convolved kernel upon Green's function.
In the literature, \cite{raissi2017machine} extended their framework by assuming observable noise. However, the assumption of LFM is too restrictive since many realistic flexible equations are nonlinear, or linear but do not have analytical Green's functions. The complete kernel is still infeasible to obtain and is hard to use expressive kernels.

To address these issues, this study models vehicle trajectories with a GP and proposes an enhanced LFM corresponding shadow GP to regularize the model. The new method is further named as PRGP \citep{wang2020physics}.
To learning the PRGP from data efficiently, we propose an inference algorithm under the posterior regularization inference framework. To justify the effectiveness of the proposed methods, we conduct a case study on freeway vehicle trajectory prediction on the NGSIM dataset \citep{colyar2007us} and compare the performance with previous influential methods.

\section{Literature review}
Microscopic traffic modeling continuously attracts research interests in the domain of transportation engineering. Over the past decades, microscopic traffic flow models, including both car-following and lane-changing models, have been widely implemented in simulation tools for evaluating the performance of various traffic operational systems. In more recent years, with the advances in communication and automation, Connected and Automated Vehicle (CAV) technologies have the potential to improve safety and efficiency, impact traffic patterns, mitigate congestion, and fundamentally change mobility. In the literature, CAV applications often require high-accuracy trajectory data extensively to support the functioning of automated driving systems. Since CAVs are expected to share the roadway network with human-driven vehicles (HVs) in a long future period, understanding and predicting HVs' trajectories also plays a key role in preventing potential traffic crashes. Hence, modeling the stochastic traffic behaviors at the microscopic level has been recognized as an important task. 

In the review of previous studies, microscopic traffic behavioral modeling can be grouped into three major categories: (a) car-following, including single-regime models \citep{pipes1953operational,forbes1963human,forbes1968driver,forbes1958measurement,chandler1958traffic, kometani1958stability,ahmed1999modeling,newell1961nonlinear,newell2002simplified}, and multi-regime models \citep{gipps1981behavioral,treiber2000congested,helbing2002micro,van1995single}; (b) lane-changing, such as rule based models (CORSIM, CTM), discrete choice models \citep{toledo2003modeling,toledo2009state, ahmed1996models, ahmed1999modeling}, stimulus response models \citep{gipps1986model,wiedemann1992microscopic,hidas2002modelling,hidas2005modelling}; and (c) overtaking \citep{jamison2007vehicle,buvrivc2009traffic,vlahogianni2012bayesian}.
Note that existing studies on studying microscopic traffic behaviors may involve one or more categories under various modeling frameworks.
The car-following models concern the safety distance \citep{gipps1981behavioral,gunay2007car}, optimal velocity \citep{bando1995dynamical,jin2010non}, rule-based celluar automata \citep{nagel1992cellular,nagel1998two}, rule-based fuzzy logic \citep{kikuchi1992car,mcdonald1997development,das1999simulations,das1999fuzzy,wu2000fuzzy,wu2003validation,ma2004toward,moridpour2012lane}, and 2-dimensional problems \citep{matcha2020simulation,delpiano2020two}. However, this study does not aim to present a historical review of car-following models, but focuses on the analytically and/or numerically differentiable closed-form formulations.
In this regard and for simplicity, car-following models can be classified by the number of equations that applied to the entire driving process into two categories: (a) single-regime models, and (b) multi-regime models. Single-regime models simplify the car-following behaviors in stable cruising situations and Pipes, Forbes, General Motors (GM), and Newell models are classical single-regime models. Their mathematical formulations are summarized below.

The Pipes model is based on a safe driving rule that the space gap should be  at least the length of a car for every ten mile per hour of traveling speed \citep{pipes1953operational}, which is formulated as follows:
\begin{equation}
    q^x_i(t)_{min} = \frac{\dot{x}_i(t)}{0.447*10}l_i
    \label{eq:pipes}
\end{equation}
where $q^x_i(t)_{min}$ refers to the minimal Spatial headway.


The Forbes model is based on the safety rule that the time gap between a vehicle and its leader should always be equal to or greater than the reaction time \citep{forbes1963human,forbes1968driver,forbes1958measurement}. The Forbes model is formulated as follows:
\begin{equation}
    q^t_i(t) = q^x_i(t) -\frac{l_i}{\dot{x}_i}\geq \bar{\tau}_i
    \label{eq:forbes}
\end{equation}
where $q^t_i(t)$ refers to the minimal temporal gaps, and $l_i$ and $\bar{\tau}$ are parameters.

The well-known Gazis-Herman-Rothery (GHR) model describes the mechanism between the vehicle and its surrounding environment \citep{chandler1958traffic, kometani1958stability,ahmed1999modeling}:
\begin{equation}
    \ddot{x}(c, t+\Delta t) = \beta_1 [\dot{x}(c, t+\Delta t)]^{\beta_2} \frac{v(c^\prime,t)-v(c,t)}{[x(c^\prime,x)-x(c,x)]^{\beta_3}-\beta_4}
    \label{eq:ghr}
\end{equation}
where $c^\prime$ is the index of the preceding vehicle and $\beta_i, i=1,\ldots,4$ are parameters.

Newell’s model hypothesizes that if a vehicle is following another vehicle on a homogeneous highway, then the time–space trajectory of the following vehicle is essentially the same as the leading vehicle with a translation in both space and time. The Newell nonlinear model is formulated as follows \citep{newell1961nonlinear}:
\begin{equation}
    \dot{x}_i(t+\Delta t_i) = \nu_i(1-e^{-\frac{\lambda_i}{\nu_i}(s-l_i)})
    \label{eq:newell1}
\end{equation}
where $s$ is the spatial headway, and $\nu_i, \lambda_i, l_i$ are parameters.Alos, the Newell linear model is formulated as \citep{newell2002simplified}:
\begin{equation}
    \dot{x}_i(t+\Delta t_i) = x_{i-1} - l_i
    \label{eq:newell2}
\end{equation}
where $l_i$ is a model parameter.

In summary, those single-regime models are mathematically concise, however, are hard to describe the various scenarios including start-up, speed-up, free flow, cut-off, following, stop and go, approaching, and stopping, etc. To address this issue, multi-regime models were proposed including the Gipps model, Intelligent Drive Model (IDM), and Van Aerde model. The formulations of a part of the models are listed as follows.

The Gipps model is formulated as\citep{gipps1981behavioral}:
\begin{equation}
    \dot{x}_i(t+\Delta t_i)=
    -b_i\tau_i+\sqrt{b^2_idt_i-b_i\dot{x}_idt_i-\frac{\dot{x}^2_{i-1}(t)}{B_{i-1}}+2l_{i-1}-2s_i(t)}
    \label{eq:gipps}
\end{equation}
where $b_i, \tau_i, B_{i-1},l_{i-1},s_i(t)$ are model parameters.

The Intelligent Drive Model (IDM) is expressed as a superposition of the follower’s acceleration term and a deceleration term which depends on the desired spacing $s^*_i$: \citep{treiber2000congested,helbing2002micro}. 
\begin{equation}
    \ddot{x}_i(t+\Delta t_i) = A_i\Big[1-(\frac{\ddot{x}_i}{\nu_i})^\delta-(\frac{s_i^*}{s_i})^2\Big]
    \label{eq:idm1}
\end{equation}
\begin{equation}
    s^*_i=s_0+s_1\sqrt{\frac{\dot{x}}{\nu_i}}+T_i\dot{x}_i+\frac{\dot{x}_i[\dot{x}_i-\dot{x}_{i-1}]}{2\sqrt{g_ib_i}}
    \label{eq:idm2}
\end{equation}
where $s_0, s_1, \nu_i, T_i, g_i,b_i,s_i, A_i, delta$ are model parameters.

The Van Aerde (VA) model combines Pipes model and Greenshields model into the following equation \citep{van1995single}:
\begin{equation}
    s_i = c_1 + c_3\dot{x}_i + c_2/(v_f - \dot{x}_i)
    \label{eq:va1}
\end{equation}
where the parameters are calculated from the macroscopic model parameters.
\begin{equation}
    c_1 = \frac{\nu_f}{k_j \nu^2_m}(2\nu_m-\nu_f)
    \label{eq:va2}
\end{equation}
\begin{equation}
    c_2=\frac{\nu_f}{k_j \nu^2_m}(\nu_f-\nu_m)^2
    \label{eq:va3}
\end{equation}
\begin{equation}
    c_3=\frac{1}{q_m} - \frac{\nu_f}{k_j \nu^2_m}
    \label{eq:va4}
\end{equation}

With the evaluation of field data, these models showed the capability of capturing car-following behaviors, however, require great efforts for parameter calibrations. Moreover, modeling under ideal theoretical conditions is difficult to handle noisy and fluctuated data collected by traffic sensors.

Given the increasing availability of high fidelity traffic data (i.e. differential GPS data), data-driven methods were developed to avoid the theoretical assumptions and reduce calibration efforts \citep{antoniou2011synthesis,zhang2011data}. 
In the literature, data-driven methods for modeling car-following behaviors can be classified into the following categories: (a) Artificial Neural Network (ANN) models \citep{yang1992neural,dougherty1992behaviourial,hunt1994modelling,dumbuya2009complexity,colombaroni2014artificial,chong2013rule,zheng2013car,zhou2017recurrent,wu2019memory}; (b) reinforcement learning or Approximate Dynamic Programming was also used to identify the behavior of the vehicle \citep{zhou2016microscopic,zhu2018human}; (c) rough set theory \citep{hao2017data}.
It should be noted that data-driven methods can be a plausible substitute for theory-based models \citep{papathanasopoulou2015towards}, however, would highly rely on the data quantity and quality. They may experience a significant accuracy drop of model performance in the following scenarios:
(i) training data are scarce and insufficient to reveal the complexity of the system; 
(ii) training data contain random noise, or include non-measurable incorrect/misleading information; and 
(iii) testing data are far from the training examples, i.e., extrapolation. 
Hence, in those models, training data are assumed to be sufficient to yield the hyper-parameters, which is unfortunately not always true in practice.

To address this issue, this paper aims to leverage a hybrid PRGP framework, developed by our pioneer work\citep{yuan2020macroscopic}, to adopt classical car-following models for enhancing the training procedure of the multivariant Gaussian process. The new method is expected to outperform both classical car-following models and existing data-driven approaches. Moreover, the difficulty of modeling lane-changing behaviors can also be overcome by the proposed model concurrently.

\section{Physics Regularized Gaussian Process}
\subsection{Motivations}
Although very few studies that hybrid traffic flow models with machine learning (ML) exist in transportation engineering, a group of research in data science proposed a promising approach that can encode physical knowledge into ML. The simpler version, named physics guided ML (PGML) \cite{jia2018physics,jia2020physics}, is based on neural network (NN) and its training objective is to minimize a new loss function that contains two terms (one is from NN that represents empirical errors and the other is from physics models that indicate physical inconsistency). Along the same logic on modifying the loss function, the advanced version, called physics informed ML (PIML) \cite{wang2017physics,wu2018physics,wang2017comprehensive,pilania2018physics,raissi2019physics,wang2019prediction,wang2017physics1,raissi2017physics}, integrates physics models with the NN family such as recurrent NN (RNN) and deep NN (DNN). Moreover, PIML can help discover the unknown model parameters that best describe the observed data. As shown in the literature, PGML and PIML were proved to be effective when dealing with a small dataset. However, both of them may suffer from the following limitations: i) they often assume the partial differential equations (PDE) are available to generate the physical inconsistency term, which prevents the adoption of discretized physics models; ii) they are based on the NN family models, which always have over-fitting problems, and their model performances still depend on the quality of the training dataset; and iii) their modeling framework is hard to work with data from multi-resources. Hence, this study aims to leverage the newly developed PRGP theory to model the stochastic nature of traffic behaviors at the microscopic level.
\subsection{Gaussian Process}
With a remarkably low computational cost in the testing phase, GP is a general framework for measuring the similarity between observations from training data to predict the unobserved values. For the convenience of discussion, key notations in this study are defined as follows.
Suppose we aim to learn a machine $\mathbf{f} :\mathbb{R}^d \rightarrow \mathbb{R}^{d^\prime}$, mapping a $d$-dimensional Euclidean space to a $d^\prime$-dimensional Euclidean space from a training set $\mathcal{D} =(\mathbf{X},\mathbf{Y})$, the input vector is denoted as $\mathbf{X}=[x_{1},\ldots,x_{N}]^\intercal$, the output vector is denoted as $\mathbf{Y}=[y_{1},\ldots,y_{N} ]^\intercal$. Here, $x$ is the $d$ dimensional input vector, and $y$ is the $d^\prime$ dimensional output vector, 
$\mathbf{f}=[f(\mathbf{x}_{1}),\ldots,f(\mathbf{x}_{N})]^\intercal$, and $\mathcal{N}(\cdot,\cdot)$ represents the Gaussian distribution.

In the GP framework, the following core assumptions are prerequisites.
The noise-free output $\mathbf{f}$ follows a multivariate Gaussian distribution $p(\mathbf{f}|\mathbf{X})=\mathcal{N}(\mathbf{f}|\mathbf{m},\mathbf{K})$ with condition on the input $\mathbf{X}$, where the kernel matrix is defined by $\mathbf{K}_{ij}=k(x_i,x_j)$ and $[\cdot]_{ij}$ denotes the element located at Row $i$, column $j$ of the matrix.
The observations $\mathbf{Y}$ have an isotropic Gaussian noise 
$p(\mathbf{Y}|\mathbf{f})=\mathcal{N}(\mathbf{f},\tau^{-1}\mathbf{I})$, where $\tau$ refers to the inverse variance, and isotropic means that the noise from each dimension is independent identically distributed (i.i.d.).
The kernel $k(\cdot,\cdot)$ is assumed to be positive-definite and smooth, where positive-definiteness means the determinant of the kernel exists, and the smoothness requires the kernel has derivatives of all orders (or at least high order) in its domain.
These weak assumptions fit most scenarios because minimal restrictions apply.

\begin{figure}[ht]
    \centering
    \includegraphics[width=0.4\textwidth]{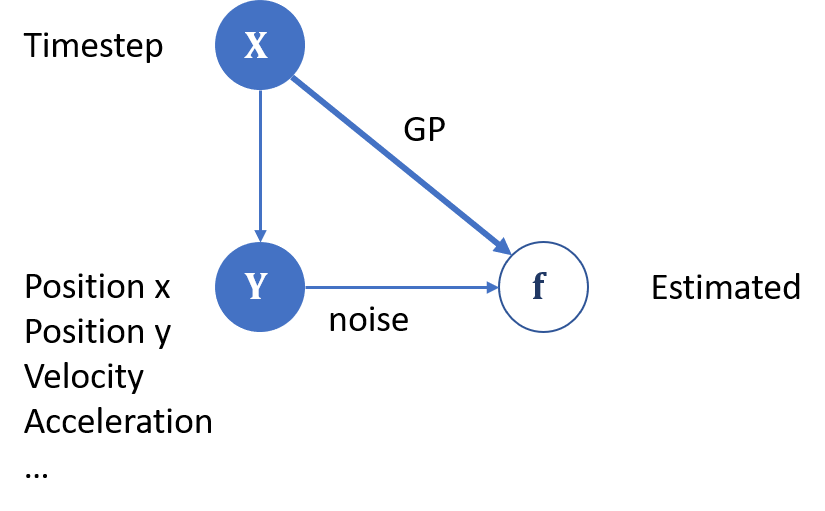}
    \caption{The conventional GP}
    \label{fig:gp}
\end{figure}

To estimate and predict the GP, we can marginalize out $\mathbf{f}$ to obtain the marginal likelihood $p(\mathbf{Y}|\mathbf{X})=\mathcal{N}(\mathbf{Y}|\mathbf{0},\mathbf{K}+\tau^{-1} \mathbf{I})$. 
The key task is to learn the kernel (i.e. covariance) function between the variables. 
Thereafter, given the new input $\mathbf{x}^*$, the $\mathbf{f}$ function value can be estimated based on Eq.~\ref{eq:pfxxy}.
\begin{equation}
    \label{eq:pfxxy}
    p(\mathbf{f}(\mathbf{x}^*)|\mathbf{x}^*,\mathbf{X},\mathbf{Y})=\mathcal{N}(\mathbf{f}(\mathbf{x}^*)|\mu(\mathbf{x}^*),\nu(\mathbf{x}^*))
\end{equation}
where the mean $ \mu(\mathbf{x}^*)=\mathbf{k}_*^\intercal (\mathbf{K}+\tau^{-1} \mathbf{I})^{-1} \mathbf{Y}$, the standard deviation $\nu(\mathbf{x}^*)=k(\mathbf{x}^*,\mathbf{x}^*)-\mathbf{k}_*^\intercal(\mathbf{K}+\tau^{-1} \mathbf{I})^{-1} \mathbf{k}_*$, and the kernel vector $\mathbf{k}_*=[k(x^*,x_1),\ldots,k(x^*,x_N)]^\intercal$ are calculated from $\mathbf{K}$.
The original GP structure in stochastic modeling is illustrated in Fig.~\ref{fig:gp}, where the circled node denotes the random tensor, the shaded node is known, and the arrows represent the conditional probability.
\subsection{Physics Regularized Gaussian Process}
Similar to other data-driven and ML models, GP highly relies on the quality of training data and its performance may below the acceptable level when the data is noisy (e.g., the obtained vehicle trajectories are often flawed). On the other hand, physical knowledge from classical traffic flow models, expressed as differential equations, provides insight into the system's mechanism and can be very useful for both estimation and prediction.
To leverage the physical knowledge in the Gaussian Process, the seminal work of \cite{alvarez2009latent,alvarez2013linear} proposed latent force models (LFM) that use convolution operations to encode physics into GP kernels.
They assume the differential equations are linear and have analytical Green's function \citep{roach1982green,stakgold2011green} which are shown below.

Denoting $\mathscr{L}$ as the linear differential operator \citep{courant2008methods} and $u(\cdot)$ as a latent force function:
\begin{equation}
    \label{eq:green1}
    \mathscr{L}=\frac{\textrm{d} }{\textrm{d}\mathbf{x}}\Big[p(\mathbf{x})\frac{\textrm{d}\mathbf{y} }{\textrm{d}\mathbf{x}}\Big]+q(\mathbf{x})
\end{equation}
where, $p(\mathbf{x})$ and $q(\mathbf{x})$ are the coefficient functions and $\mathbf{y}$ is an unknown function of the free variable $\mathbf{x}$, the vector-valued boundary condition operator $\Vec{D}$ can be expressed as:
\begin{equation}
    \label{eq:green2}
    \Vec{D}u=\begin{bmatrix}
        \alpha_1u^\prime(0)+\beta_1u(0)\\
        \alpha_2u^\prime(l)+\beta_2u(l)\\
    \end{bmatrix}
\end{equation}
where, $\alpha_1$, $\alpha_2$, $\beta_1$, and $\beta_2$ are weighting parameters.
Also, let $f(x)$ be a continuous output function in $[0,l]$, then there is one and only one solution $u(x)$ that satisfies:
\begin{equation}
    \label{eq:green3}
    \mathscr{L}u=f
\end{equation}
\begin{equation}
    \label{eq:green4}
    \Vec{D}u=\Vec{0}
\end{equation}
and it is given by:
\begin{equation}
    \label{eq:green5}
    u(x)=\int f(s)\mathcal{G}(x,s)ds
\end{equation}
where $\mathcal{G}(x,s)$ is a Green's function.

Given this assumption, the kernel of the target function can be derived by convolving the Green's function with the kernel of the latent functions.
LFM considers $W$ output functions ${f_1(x),\ldots,f_w(x),\ldots,f_W(x)}$, and assumes each output function $f_w$ is governed by a linear differential equation.
\begin{equation}
    \label{eq:lfm}
    \mathscr{L}f_{w}(\mathbf{x}) = u_{w}(\mathbf{x})
\end{equation}

The latent force function $u$ can be further decomposed as a linear combination of several common latent force functions as follows.
\begin{equation}
    \label{eq:udecomp}
    u_w(\mathbf{x})=\sum_{r=1}^{R}s_{rw} g_{r}(\mathbf{x})
\end{equation}
where $R$ is the number of decomposed force functions, $s$ is the latent matrix.
Since $\mathscr{L}$ is linear, if we assign a GP prior over $u(x)$, $f_w(x)$ has a GP prior as well.
Moreover, if the Green's function, namely the solution of Eq.~\ref{eq:green}, is available, we can obtain Eq.~\ref{eq:green_sol}.
\begin{equation}
    \label{eq:green}
    \mathscr{L}\mathcal{G}(\mathbf{x},\mathbf{s})=\delta(\mathbf{s}-\mathbf{x})
\end{equation}
where $\delta$ is the Dirac delta function.  Green's function of linear operator $\mathscr{L}$ over the Euclidean space $\mathbb{R}^d$ is defined as the solution of Eq.~\ref{eq:green}.
\begin{equation}
    \label{eq:green_sol}
    f_w(x)=\int \mathcal{G}(x,s)u_{i}(\mathbf{s})\textrm{d}\mathbf{s}
\end{equation}

Hence, given the kernel for $u_w$, we can derive the kernel for $f_w$ through a convolution operation which is shown in Eq.~\ref{eq:fi_kernel}.
\begin{equation}
    \label{eq:fi_kernel}
    k_{f_w}(\mathbf{x}_1,\mathbf{x}_2)=\iint \mathcal{G}(\mathbf{x}_1,\mathbf{s}_1)\mathcal{G}(\mathbf{x}_2,\mathbf{s}_2)k_{u_w}(\mathbf{s}_1,\mathbf{s}_2)\textrm{d}\mathbf{s}_1\textrm{d}\mathbf{s}_2
\end{equation}

The LFM has been applied to several linear differential models, such as a biological network motif model (first-order dynamical system), a linearized human motion model (second-order dynamical system), and a heavy metal pollutant diffusion model (a partial differential equation).
The biological network motif model is formulated in Eq.~\ref{eq:sample1}.
\begin{equation}\label{eq:sample1}
    \frac{\textrm{d} \mathbf{y}_q}{\textrm{d}\mathbf{x}} + \alpha_q \mathbf{y}_q = \beta_q + \sum^{R}_{r=1}S_{rq}f_r(\mathbf{x})
\end{equation}
where $\alpha_q, \beta_q$ are biological parameters.
    
Solving Eq.~\ref{eq:sample1} for $\mathbf{y}_q$, we obtain Eq.~\ref{eq:sample2}.
\begin{equation}\label{eq:sample2}
    \mathbf{f}_{rq} = s_{rq}\exp(-\alpha \mathbf{x})\int^\mathbf{x}_0f_r(\mathbf{\chi})\exp(\alpha_q\chi)\textrm{d}\chi
\end{equation}
If each latent force is taken to be independent with a kernel function, the output kernel function can be computed analytically \citep{lawrence2007modelling}. Then, the cross-covariance between the inputs and outputs can be computed analytically.

In the case of second order dynamic system, the human motion is represented by a skeleton and multivariate time courses of angles. The original model of the motion in Eq.~\ref{eq:sample3} is nonlinear due to the variations of the centers of mass, and was linearized to apply LFM.
\begin{equation}\label{eq:sample3}
    \frac{\textrm{d}^2\mathbf{y}_q(\mathbf{x})}{\textrm{d} \mathbf{x}^2} + \alpha_q \frac{\textrm{d} \mathbf{y}_q(\mathbf{x})}{\textrm{d} \mathbf{x}} + \beta_q \mathbf{y}_q = \phi_q + \sum^R_{r=1}s_{rq}f_r({\mathbf{x}})
\end{equation}
where $\alpha_q, \beta_q, \phi_q$ are physical parameters.
Following the similar procedure, the second order model is solved in Eq.~\ref{eq:sample4}.
\begin{equation}\label{eq:sample4}
    \mathbf{f}(\mathbf{x}) = \frac{s_{rq}}{\sqrt{4\beta_q-\alpha_q^2}/2}\exp (-\frac{\alpha_q}{2}\mathbf{x}) \times \int^{\mathbf{x}}_0f_r(\chi)\exp (\frac{\alpha_q}{2}\chi)\sin (\sqrt{\beta_q-\alpha_q^2}/2\cdot(\mathbf{x}-\chi))\textrm{d}\chi
\end{equation}
Then, the covariance between the inputs and cross-covariance between the inputs and outputs can be computed analytically.

In the third case, the partial differential model was presented in Eq.~\ref{eq:sample5}.
\begin{equation}\label{eq:sample5}
    \frac{\partial y(x,t)}{\partial t}=\sum^{J}_{j=1}\kappa_q \frac{\partial^2 y_q(x,t)}{\partial x^2}
\end{equation}
where the independent variable $(x,t)$ is 2-dimensional, $\kappa_q$ is the physical parameter. The solution was given in Eq.~\ref{eq:sample6} \citep{polyanin2015handbook}.
\begin{equation}\label{eq:sample6}
    \mathbf{y}_q(x,t)=\sum^R_{r=1}s_{rq}\int_{\mathbb{R}^d}f_r(x^\prime)G_q(x,x^\prime,t)\textrm{d} x^\prime
\end{equation}
where the specific Green function is given by Eq.~\ref{eq:sample7}.
\begin{equation}\label{eq:sample7}
    G_q(x,x^\prime,t)=\frac{1}{2^d\pi^{d/2}(\kappa_qt)^{d/2})} \exp \Big[ -\sum^d_{j=1}\frac{(x_j-x^\prime_j)^2}{4\kappa_qt}\Big]
\end{equation}

With the analytical solutions, \cite{alvarez2013linear} proposed the deterministic training conditional variational approximation (DTCVAR) to use the lower bound on the marginal likelihood to yield the parameters. 
To enable the kernel convolution, LFM requires that the differential equations must be linear and have analytical Green's functions.
However, many realistic differential equations are either nonlinear or linear but do not possess analytical Green's functions, hence they cannot apply the LFM framework.

In some other cases, even with a tractable Green's function, the complete kernel of all the input variables is still infeasible to obtain.
To obtain an analytical kernel after the convolution, we have to convolve Green's functions with smooth kernels.
This may prevent us from integrating the physical knowledge into more complex yet highly flexible kernels, such as deep kernel \citep{wilson2016deep}.
To handle the intractable integral, we need to develop extra approximation methods, such as Monte-Carlo approximation.

Given the differential equations that describes the physical knowledge in Eq.~\ref{eq:pk}, the proposed augmented LFM equation is formulated in Eq.~\ref{eq:nonlinearlfm}.
\begin{equation}
    \label{eq:pk}
    \Psi f(\mathbf{x}) = 0
\end{equation}
\begin{equation}
    \label{eq:nonlinearlfm}
    \Psi f(\mathbf{x}) = g(\mathbf{x})
\end{equation}
where the differential operator $\Psi$ can be linear, nonlinear, or numerical differential operator, $g(\cdot)$ represents the unknown latent force functions, $f(\mathbf{x})$ is the function to be estimated from data $\mathcal{D}$.
We aim to create a generative component to regularize the original GP with a differential equation.
Using Augmented LFM, the differential equation is encoded to another GP, which is called a shadow GP.
To yield the numerical outputs, the kernel of the shadow GP should be efficiently learnable.

\begin{figure}[H]
        \centering
        \includegraphics[width=0.8\textwidth]{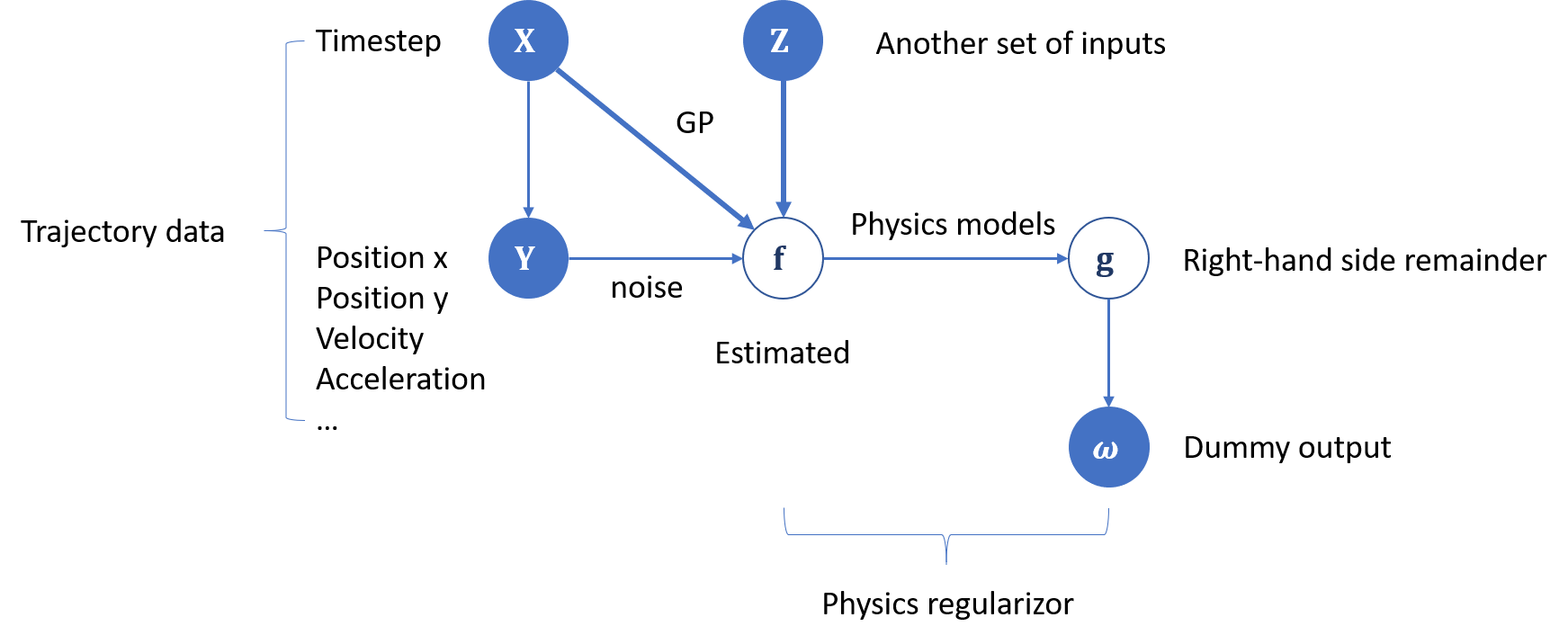}
        \caption{The stochastic model of the proposed PRGP}
        \label{fig:PRGP}
\end{figure}

Note that $f(\mathbf{x})$ in Eq.~\ref{eq:pk} is the static-value function and $g(\mathbf{x})$ in Eq.~\ref{eq:nonlinearlfm} are assumed to be the stochastic variables following a GP.
More specifically, Fig.~\ref{fig:PRGP} illustrates the structure of the proposed PRGP.
The difference from the conventional GP is that the PRGP employs the pseudo observations $(\mathbf{Z}, \omega)$ to encode the physical knowledge in the differential equation in Eq.~\ref{eq:pk}.
Also, \cite{heinonen2018learning} pointed out the numerical integral mapping (namely, the forward method) is computationally intense, and the gradient mapping approximates the differential equation (namely, the inverse method) relatively efficiently \citep{ramsay2007parameter}.
To conduct the inference and prediction, the resultant PRGP is still known to be a GP, which is proven in \ref{app:post_PRGP}.

To deal with the multiple outputs, we can place independent GP priors over common latent function $g_r$, then each $u_w$ and $f_w$ will obtain GP priors in turn.
Via a similar convolution, we can derive the kernel across different outputs (i.e. cross-covariance) $k_{f_w,f_{i'}}$.
In this way, the physical knowledge in the Green's function is hybridized with the kernel for the latent forces.
This procedure is used to learn the GP model with a convolved kernel from the training data.
To apply the proposed method with multiple differential equations of microscopic behaviors, the extended structure of PRGP is presented. The multi-equation and multi-output framework of applying the proposed method to capture the stochastic traffic behavior process in the car-following model.

In summary, given a collection of training data $\mathcal{D} = (\mathbf{X}, \mathbf{Y})$, the training objective of the proposed PRGP is of the following form, 
\begin{align}
    C = M(\mathbf{Y}|\mathbf{X}) + \lambda \cdot R\big(p\left(f(\cdot)|\mathcal{D}\right) \big) \label{eq:obj}
\end{align}
where $M(\mathbf{Y}|\mathbf{X})$ is the log model evidence (i.e., log marginal likelihood of the outputs given the inputs) and $R(\cdot)$ is a regularization term that regularizes the posterior of the target function $p\big(f(\cdot)|\mathcal{D})$ with the physics encoded in the differential equation $\psi f(x) = g(x)$.

\section{Stochastic Modeling of Microscopic Traffic Behavior}
\subsection{Critical Issues}
\begin{figure}[H]
        \centering
        \includegraphics[width=0.8\textwidth]{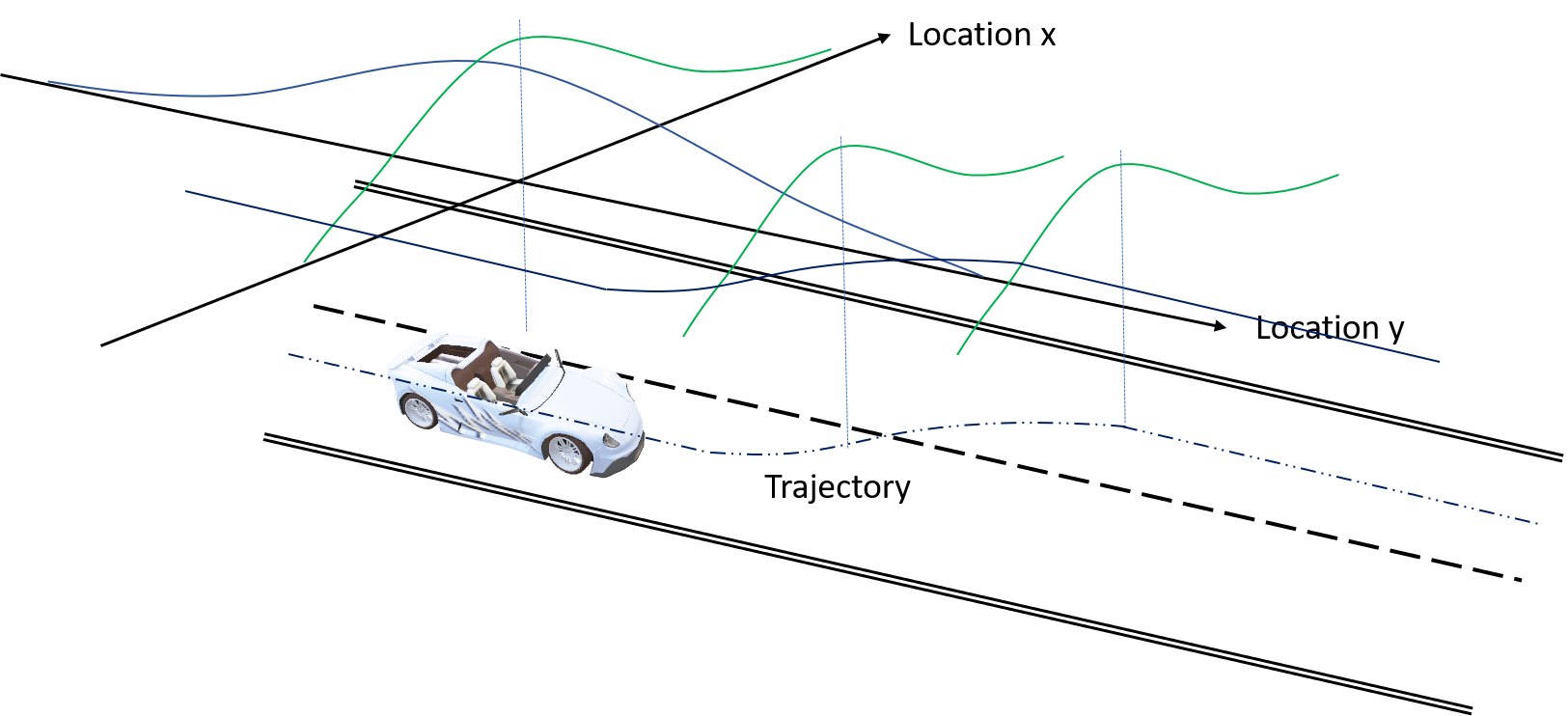}
        \caption{Modeling vehicle trajectory with Gaussian Process}
        \label{fig:traj}
\end{figure}

Notably, lane-changing is a discrete event and its models cannot be encoded into PRGP with the car-following models. Hence, one of the main contributions of the proposed modeling framework is that we can use the GP part to predict lane-changing activities. The estimation and prediction of GP are linear algebra operations and is computationally efficient.
Fig.~\ref{fig:traj} shows the GP in the local coordinates, where the green curve is the distribution of the location in $y$ axis, and the blue curve is the distribution of the location in $x$ axis. The local coordinates are assumed to be GP. It means the coordinates of the vehicle in each time step follow Gaussian distribution, respectively. The x coordinates in the trajectory are correlated, and the correlation function can be captured by a smooth positive definitive kernel function. The kernel function depends on the time step input and a few learnable hyper-parameters. Then, the velocity, acceleration, preceding vehicle velocity, and occurring of lane-changing can be easily obtained by analyzing $x$ and $y$ over time $t$.  

\subsection{Stochastic Car-following Models}
To replicate the realistic traffic behavior, car-following forms one of the main processes in all microscopic simulation models as well as in modern traffic flow theory, which attempts to understand the interplay between phenomena at the individual driver level and global behavior on a more macroscopic scale \citep{brackstone1999car}.
Based on the physical knowledge, the stochastic car-following behaviors can be modeled as differential equations, where the target is the tuple of the vehicle status $(\mathbf{x},\mathbf{\dot{x}},\mathbf{\ddot{x}})$ in the system given the vehicle $c$ at time $t$.
By the definition, the velocity $\dot{x}$ and the acceleration $\ddot{x}$ are formulated in the following equations.
\begin{equation}
    \mathbf{\dot{x}} = \frac{\partial \mathbf{X}}{\partial t}
    \label{eq:velocity}
\end{equation}
\begin{equation}
    \mathbf{\ddot{x}} = \frac{\partial \mathbf{\dot{X}}}{\partial t}
    \label{eq:acc}
\end{equation}

In the microscopic traffic streams, the trajectory of a vehicle $c$ is impacted by its surrounding environment, such as the road geometry and nearby vehicles.
Assuming the feasible longitude position of the vehicle is within the interval $[x_{lngmin}, x_{lngmax}]$, the longitude position constraint is formulated in Eq.~\ref{eq:road}.
\begin{equation}
    x_{lngmin} \leq [x(c, t)]_{lng} \leq x_{lngmax} \quad \forall c, t
    \label{eq:road}
\end{equation}

The following equation shows the formula of $c^\prime$, given the vehicle $c$, the time $t$, and the set of vehicles $C$. \cite{papathanasopoulou2019towards} formulated a 2-dimensional leader-follower identification model.
In this paper, the preceding vehicle $c^\prime$ is defined as the first vehicle in front of the vehicle $c$ in the data processing.
\begin{equation}
    c^\prime= \arg\min_{\bar{c}\in C\setminus \{c\}} \Vert[x(c, t)]_{lng} - [x(\bar{c}, t)]_{lng}\Vert
    \label{eq:cons_c1}
\end{equation}
s.t.
\begin{equation}
    \Vert\left [x(c, t)]_{hor} - [x(c^\prime, t)]_{hor}\right \Vert \leq \xi
    \label{eq:cons_c2}
\end{equation}

\begin{equation}
    [x(c, t)]_{lng} - [x(c^\prime, t)]_{lng} > \delta
    \label{eq:cons_c3}
\end{equation}
where the index $lng$ denotes the longitudinal dimension, the index $hor$ represents the horizontal dimension, $\xi$ is a horizontal distance threshold for two vehicles in the same lane, and $\delta$ is the minimal space headway.

To compare the performances of encoding different car-following models into PRGP, this study tests 7 widely implemented models including the Pipes model, the Forbes model, the GHR model, the Newell nonlinear model, the Newell linear model, the Gipps model, and the VA model. Based on their model formulations, the corresponding partial differential equations that represent the physical knowledge can be obtained by the  equations shown in the following table.
\begin{table}[H]
    \centering
    \caption{Summary of the partial differential equations of different car-following models}
    \begin{tabular}{p{4cm} p{12cm}}
    \toprule
        Model  & Partial Differential Equation\\
    \midrule
        Pipes      & \begin{equation}
   \Psi f(c, t) = q^x_i(t) - \dot{x}_i(t) * \beta_0 = g \label{eq:pipes_prgp}
    \end{equation}       \\ 
    Forbes      & \begin{equation} \Psi f(c, t) = q^t_i(t) - q^x_i(t) + \dot{x}_i*\beta_0 = g \label{eq:forbes_prgp} \end{equation}       \\
    GHR      & \begin{equation} \Psi f(c, t) = \ddot{x}(c, t+\textrm{d}t) - \beta_1 [\dot{x}(c, t+\textrm{d}t)]^{\beta_2}=g \label{eq:ghr_prgp} \end{equation}    \\
    Newell nonlinear      & \begin{equation} \Psi f(c, t) = \dot{x}_i(t+\textrm{d}t_i) - \nu_i(1-e^{-\frac{\lambda_i}{\nu_i}(s-l_i)}) = g \label{eq:newell1_prgp} \end{equation}       \\
    Newell linear      & \begin{equation} \Psi f(c, t) = \dot{x}_i(t+\textrm{d}  t_i) - x_{i-1} + l_i = g \label{eq:newell2_prgp} \end{equation}       \\
    Gipps      & \begin{equation} \Psi f(c, t) = \dot{x}_i(t+\textrm{d} t_i) + \beta_0-\sqrt{\beta_1+\beta_2\dot{x}^2_{i-1}(t)-2s_i(t)} = g \label{eq:gipps_prgp} \end{equation}       \\
    VA      & \begin{equation} \Psi f(c, t) = s_i - \beta_0 - \beta_1\dot{x}_i - \beta_2/(\beta_3 - \dot{x}_i) = g \label{eq:va_prgp}  \end{equation}      \\
    \bottomrule
    \end{tabular}
    \label{tab:abbr}
\end{table}
More specifically, the equations of car-following models are converted to the partial differential form of Eq.~\ref{eq:nonlinearlfm} to create a stochastic model $\mathbf{f}: (c,t) \rightarrow (x,\dot{x},\ddot{x})$. 

\subsection{Posterior Regularization Inference}
Given the complexity of the proposed PRGP, a posterior regularization method is used to estimate the parameters and ensure computational efficiency.
Posterior regularization is a powerful inference methodology in the Bayesian stochastic modeling framework \citep{ganchev2010posterior}.
The physical knowledge or constraints are considered by the penalty term in posteriors, rather than through the priors and a complex intermediate computing procedure.
Based on the posterior regularization inference framework, an efficient stochastic optimization algorithm can be developed to maximize the evidence lowerbound (ELBO) of the system likelihood, as shown in Eq.~\ref{eq:lllhlb_traffic}.
The objective includes the model likelihood on data and a penalty term that encodes the constrains over the posterior of the latent variables.
\begin{equation}
    \label{eq:lllhlb_traffic}
    \begin{split}
        \log [p(\mathbf{Y},\mathbf{\omega}|\mathbf{X})]\geq \mathcal{L}=& \sum_{i=1}^{d^\prime} \log[\mathcal{N}([\mathbf{Y}]_i|\mathbf{\omega},\hat{\mathbf{K}}_i+\tau^{-1}\mathbf{I})]\\
        & +\sum_{w=1}^W\Gamma_w \mathbb{E}_{p(\hat{\mathbf{f}}_w|\mathbf{Z},\mathbf{X},\mathbf{Y})} [\log [\mathcal{N}(\Psi \hat{\mathbf{f}}_w|\omega,\hat{\mathbf{K}}_w)]]
    \end{split}
\end{equation}

Given the expectation out of the logarithm, the ELBO is numerically differentiable.
To find the numerical differentiation of ELBO, existing automatic differentiation methods are leveraged \citep{zhang2019quantifying}.
Then we can maximize $\mathcal{L}$ using the stochastic optimization shown in Algorithm~\ref{alg:1}.\\

\begin{algorithm}[H]
    \caption{The stochastic inference algorithm}
    \label{alg:1}
    \SetAlgoLined
    \KwResult{Learned kernel parameters}
     Initialization\;
     \While{not reach stopping criteria}{
        Sample a set of input locations $\mathbf{Z}$\;
        Predict the mean $\mu$ and the variance $\nu$ of $\mathbf{f}$;
        Generate a parameterized sample of the posterior target function values $\hat{\mathbf{f}}$ by the reparameterization in $f=\mu+\nu*\epsilon, \epsilon=\mathcal{N}(0,1)$\;
        Substitute the parameterized samples $\hat{\mathbf{f}}$ to obtain the unbiased predicted ELBO $\tilde{\mathcal{L}}$ in Eq.~\ref{eq:lllhlb_traffic}\;
        Calculate $\nabla_\theta\tilde{\mathcal{L}}$, an unbiased stochastic gradient of $\tilde{\mathcal{L}}$ via the auto-differential technique\;
        Update the parameters $\theta$ via the gradient decent in $\theta^{t+1}=\theta^{t}+\alpha\nabla_\theta\tilde{\mathcal{L}}$ ;
     }
\end{algorithm}
where $\alpha$ refers to the learning rate and $\theta$ denotes all trainable parameters.\\

The correctness of Algorithm~\ref{alg:1} is proven in \ref{app:alg} and the computational complexity of the proposed inference algorithm is analyzed as follows.
The time complexities of the inference of the original GP and the shadow GP are $O(N^3)$ and $O(m^3)$, respectively.
Thus, the total time complexity for the inference of the two GPs is $O((Nd^\prime)^3+m^3)$.
To store the kernel matrices of the original GP and the shadow GP, the space complexity is $O((Nd^\prime)^2+m^2)$.
In the testing phase, the time complexity of the model prediction is marginal (e.g., less than $1$ ms) empirically.

\subsection{Algorithm Overview and Comparison}
\begin{figure}[H]
    \centering
    \includegraphics[width=0.8\textwidth]{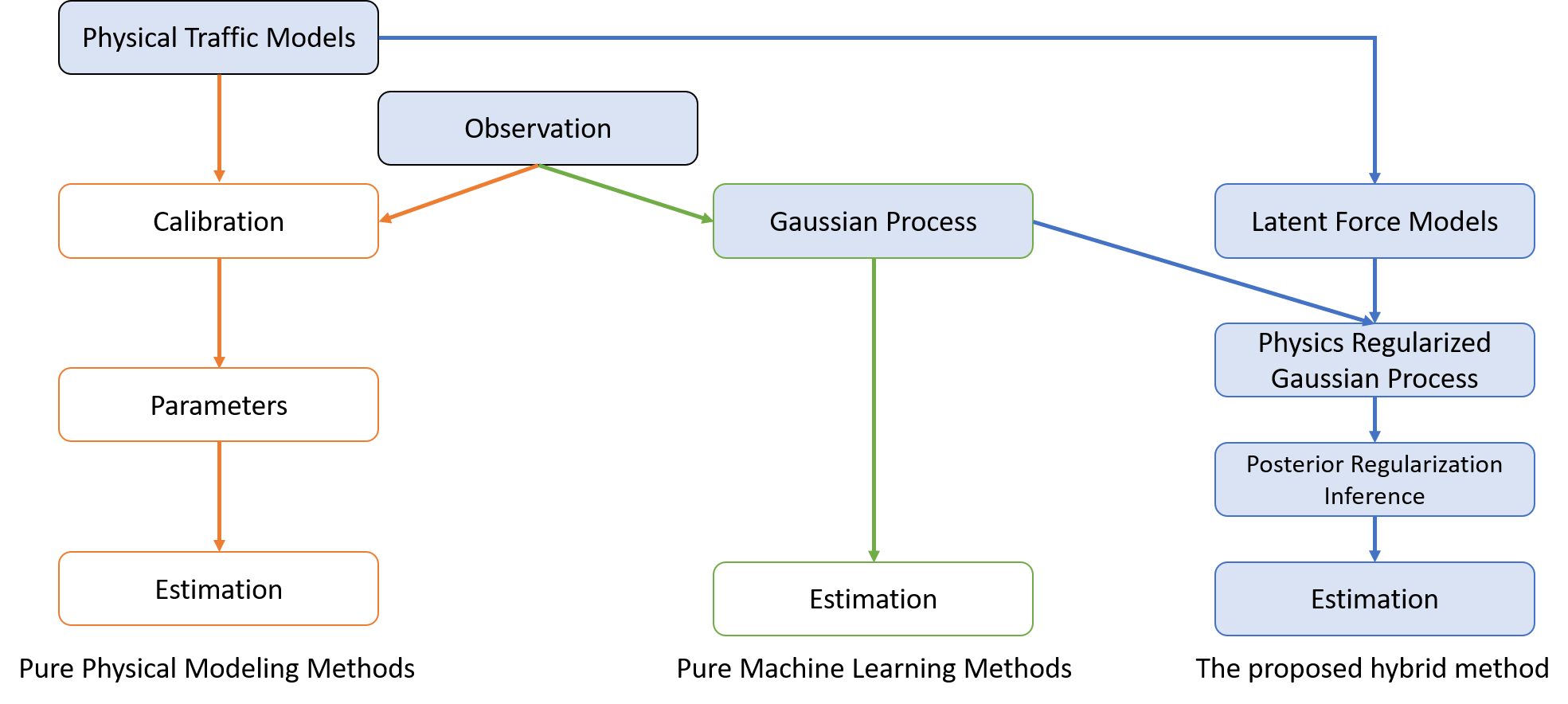}
    \caption{The algorithm overview of the proposed method and the comparable methods}
    \label{fig:overview}
\end{figure}

In summary, Fig.~\ref{fig:overview} compares the proposed hybrid method with the conventional physics model (i.e., car-following) and the pure machine learning method (i.e., GP), where the pure physics modeling workflow is in orange, the pure machine learning workflow is in green, and the workflow of the proposed PRGP is in blue. By encoding the physics model as a shadow GP and integrating it with the original GP, the new training objective as shown in Eq. \ref{eq:obj} will be created. Then the valuable knowledge from the physics model can function as a regularizer in the model training process. Therefore, the proposed PRGP-based model is expected to outperform conventional traffic flow models in capturing estimation uncertainties and outperform pure machine learning models in dealing with small and noisy data.  

\section{Case study}
\subsection{Scenario setup}
This study leverages the open Next Generation SIMulation (NGSIM) data \citep{alexiadis2004next} to justify the effectiveness of the proposed model.
The partial fields of the original NGSIM data include time, vehicle ID, location x, location y, velocity, acceleration, preceding vehicle ID, space headway, and time headway. Note that the locations are of local relative coordination. To yield the preceding vehicle velocity, a prepossessing procedure is employed to match the preceding vehicle ID at the same time. The original time is preprocessed to the time difference between the starting time and the time value in seconds. The output variable representation $\mathbf{y}=(position_x, position_y, velocity, acceleration, preceding~velocity, space, time~headway)$, where the $position_x, position_y$ refer to the local coordinates.
The input variables include the time of each record. In the literature, the data index representation has variations, such as (vehicle label, time interval), (vehicle label, week, day-of-week, time interval). In the experiments, we use the compatible representation (time interval), namely $(k)$, for the consistence purpose. 
Note that the other variations of structural representation of the data are fully compatible with the proposed framework, and the impact of the data representation may depend on the specific case.

In the cases, the data is randomly shuffled and split into the training set and testing set separately. The training and calibration of the methods are based on the training dataset. The separated testing dataset, collected from the same location and time period, is used for a cross-validated fair comparison. 
Note that the inputs of the proposed PRGP based method and the microscopic traffic model are different, as the latter method requires more detailed observations. The proposed method assumes the unobserved disturbance or accidents in the framework and does not require such data.

In the experiments, the parameters of the proposed method are set as follows: (a) the number of pseudo observations $m=10$.
The proposed inference algorithm is implemented in the Tensorflow framework, where
the optimizer \emph{ADAM} is chosen as the parameter optimizer to find the negative ELBO \citep{kingma2014adam}.



The methods are designed to capture the vehicle's 2-dimensional position, velocity and acceleration, and also output auxiliary variables, such as the velocity of the preceding vehicle and the Spatial and temporal headway, at each time step.
To clarify the the equations used in each model, the list of models to be tested and compared are summarized in Table~\ref{tab:abbr2}.
\begin{table}[H]
    \centering
    \caption{Method settings in the case study}
    \begin{tabular}{llc}
    \toprule
        Abbreviation  & Model  & Equation Indices\\
    \midrule
        GP       & Multi-output Gaussian Process        & \ref{eq:pfxxy}\\
        PRGP-DEF & PRGP with velocity and acceleration definition & \ref{eq:velocity},\ref{eq:acc}\\
        PRGP-Pipes  & PRGP with Pipes model & \ref{eq:pipes_prgp}\\
        PRGP-Forbes & PRGP with Forbes model&\ref{eq:forbes_prgp}\\
        PRGP-GHR    &PRGP with GHR model&\ref{eq:ghr_prgp}\\
        PRGP-Gipps  &PRGP with Gipps model&\ref{eq:gipps_prgp}\\
        PRGP-NN     &PRGP with nonlinear Newell model&\ref{eq:newell1_prgp}\\
        PRGP-NL     &PRGP with linear Newell model&\ref{eq:newell2_prgp}\\
        PRGP-VA     &PRGP with Van Aerde model&\ref{eq:va_prgp}\\
    \bottomrule
    \end{tabular}
    \label{tab:abbr2}
\end{table}

To quantify the precision of outputs, Rooted Mean Squared Error (RMSE) and Mean Absolute Percentage Error (MAPE) of each dimension are used as the performance metric, which are defined in Eqs.~\ref{eq:def_rmse}-\ref{eq:def_mape}. Those measures of goodness-of-fit have been widely used in the previous studies \citep{antoniou2013dynamic,papathanasopoulou2015towards}: 
\begin{equation}
RMSE_j = \sqrt{\frac{1}{N}\sum_{i=1}^{N}{(\frac{||[\mathbf{y}_j]_{i}-[\hat{\mathbf{f}}_j]_i||}{\sigma_i})^2}}, \forall j\in {1,\ldots,d^\prime}
\label{eq:def_rmse}
\end{equation}
\begin{equation}
MAPE_j = \frac{100\%}{n}\sum_{i=1}^{n}{\Big\vert\frac{[\mathbf{y}_j]_{i}-[\hat{\mathbf{f}}_j]_i}{[\mathbf{y}_j]_{i}}\Big\vert}, \forall j\in {1,\ldots,d^\prime}
\label{eq:def_mape}
\end{equation}

\begin{figure}[H]
    \centering
    \includegraphics[height=6cm]{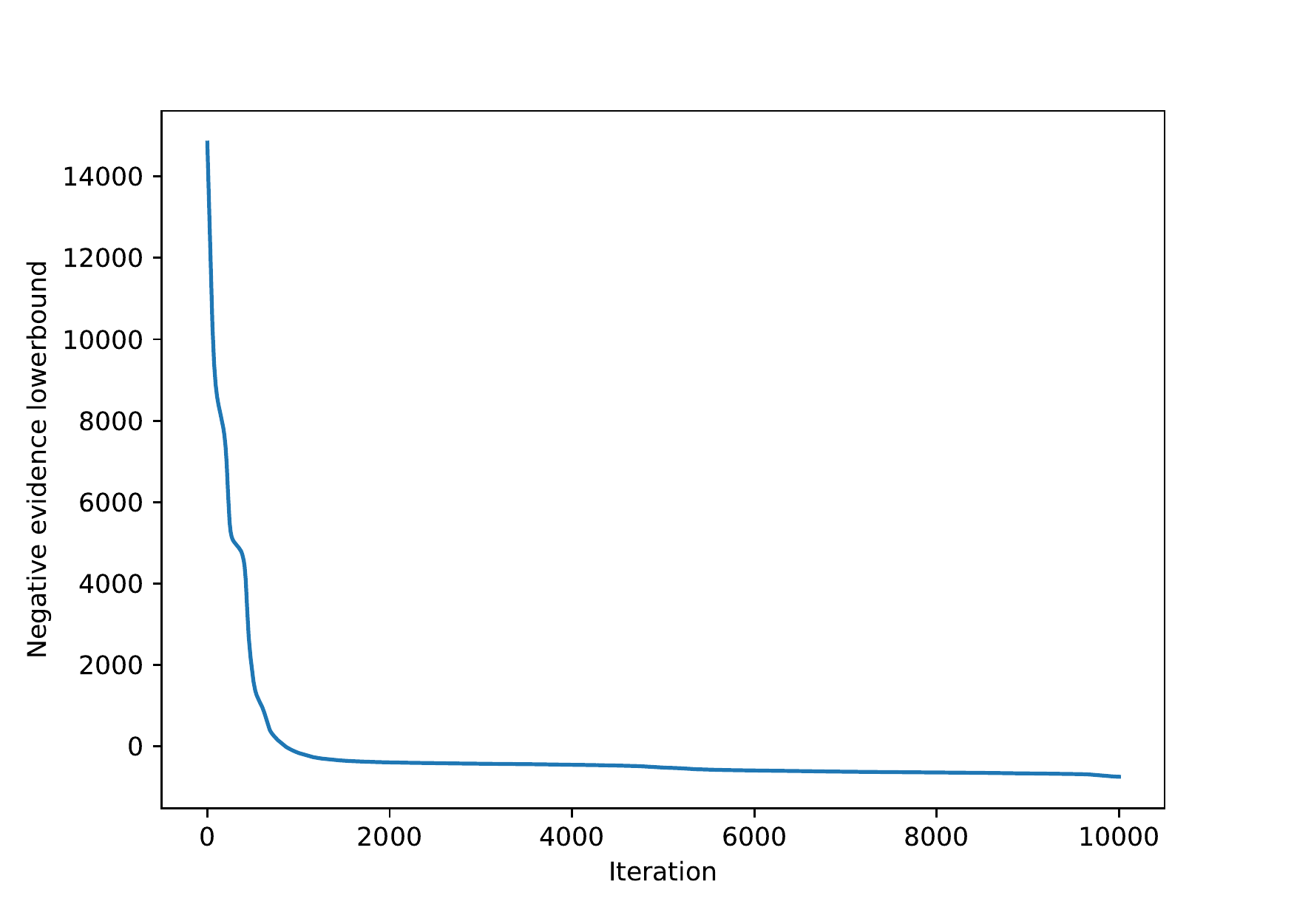}
    \caption{The convergence of the negative evidence lowerbound of log likelihood}
    \label{fig:res_elbo}
\end{figure}

The training process of $10,000$ iterations and $500$ samples costs $4,280$ seconds on a workstation equipped with a 3.5GHz 6-core CPU.
In the testing phase, the time complexity of the model prediction is marginal (e.g., less than $1$ second) empirically.
The computational process can be accelerated by about $5$ time by NVIDIA CUDA-capable GPU.
Taking the PRGP-Pipes as an example, the convergence of the proposed method is shown in Figs.~\ref{fig:res_elbo}. 

\subsection{Result analysis of Case I}
\begin{figure}[H]
    \centering
    \includegraphics[width=0.7\textwidth]{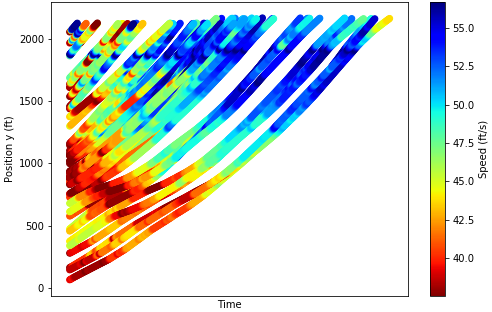}
    \caption{The vehicle trajectory data in the studied case}
    \label{fig:data3}
\end{figure}

In Case I, the proposed PRGP models are tested on the sub-dataset that contains no lane-changing records. The time range of the data is from 5:00 PM to 5:15 PM. Fig.~\ref{fig:data3} shows the trajectory data along with the speed distributions from NGSIM that are used for testing, where the trajectories of $50$ vehicles are used for training and testing. Each curve represents the continuous trajectory of one vehicle and the color labels of the curves represent the corresponding time-dependent speeds (in feet per second). 


\subsubsection{Comparison with physics models and the GP}
To validate the effectiveness of the proposed PRGP models, this study further compare their model performances on vehicle trajectory estimations with both physics (car-following) models and pure GP model. Herein, the physics models, including velocity and acceleration definitions (vel-DEF and acc-DEF), Pipes, Forbes, GHR, Gipps, VA, Newell nonlinear (NN), and Newell linear (NL), are tested as the baselines. Notably, vel-DEF and acc-DEF models are calculated from the data numerically in Eqs.~\ref{eq:velocity2}-\ref{eq:acc2}, where $\Delta t$ is the time gap between two records.
\begin{equation}
    velocity = \frac{\partial position_y}{\partial t} = \frac{position_y(t+\Delta t)}{\Delta t}
    \label{eq:velocity2}
\end{equation}
\begin{equation}
    acceleration = \frac{\partial velocity}{\partial t} =\frac{ velocity(t+\Delta t)}{\Delta t}
    \label{eq:acc2}
\end{equation}

Recognizing the importance of calibrating model parameters before implementations, this paper follows the procedure introduced in the previous studies \citep{treiber2014microscopic,fang2014calibration,cunto2008calibration,kesting2008calibrating,chen2010calibration}. The results of the calibrated parameters in each physics models are presented in Table~\ref{tab:pk_para1}, where the predicated quantity is the outputted variable from the model. As shown in Table~\ref{tab:physics}, the predicated quantities for the models include velocity, acceleration, and space gap. The RMSE of all predicated quantities are within the acceptable range, while the MAPE of the acceleration is relatively high for acc\_DEF and GHR models. The main reason is that the value of acceleration rates have a lower order of magnitude compared with others. 

\begin{table}[H]
    \centering
    \caption{The results of calibrated physics models in Case I}
    \begin{tabular}{cc}
    \toprule
    Model & Parameter\\
    \midrule
    Pipes & $\beta_1=3.6$\\
    Forbes & $\beta_1=0.81$\\
    GHR & $\beta_1= 0.8,\beta_2=2.0,\beta_1=1.5$\\
    Gipps & $b = 1,t = 0.1,B = 1,l = 6$\\
    NL& $v_i=40,\lambda=2.49, l_i=33.16$\\
    VA& $v_f = 11.11,k_j = 0.25,v_m = 8.33,q_m = 0.708$\\
    \bottomrule
    \end{tabular}
    \label{tab:pk_para1}
\end{table}

\begin{table}[H]
    \centering
    \caption{The calibration performance of the physics models in Case I}
    \begin{tabular}{cccc}
        \toprule
        Model & Predicated quantity & RMSE & MAPE \\
        \midrule
        Vel-DEF        & velocity      & 0.346     & 0.484\\
        Acc-DEF    & acceleration  & 1.588     & 135.44\\
        Pipes               & space gap     & 2.147     & 1.708\\
        Forbes              & space gap     & 1.996     & 1.667\\
        GHR                 & acceleration  & 5.20      & 370.85\\
        Gipps               & velocity      & 2.735     & 5.023\\
        NL                  & velocity      & 3.825     & 7.187\\
        NN                  & velocity      & 3.334     & 5.709\\
        VA                  & space gap     & 4.840      & 4.760\\
        \bottomrule
    \end{tabular}
    \label{tab:physics}
\end{table}

\begin{figure}[H]
    \centering
    \begin{subfigure}[b]{0.3\textwidth}
        \includegraphics[width=\textwidth]{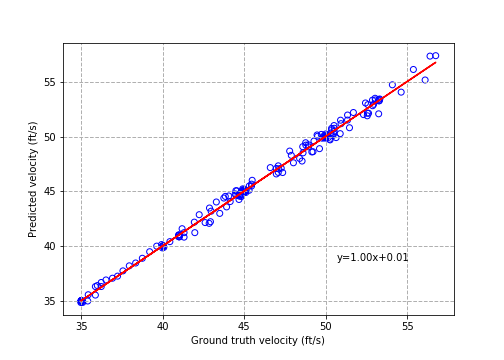}
        \caption{Vel-DEF}
    \end{subfigure}
        \hfill
    \begin{subfigure}[b]{0.3\textwidth}
        \includegraphics[width=\textwidth]{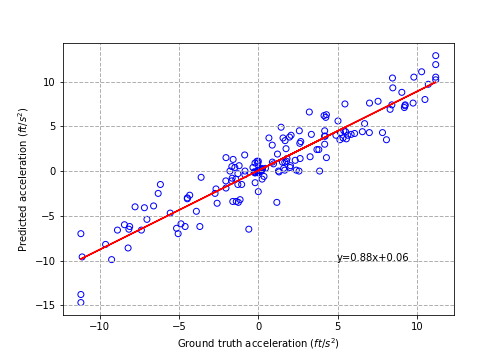}
        \caption{Acc-DEF}
    \end{subfigure}
        \hfill
    \begin{subfigure}[b]{0.3\textwidth}
        \includegraphics[width=\textwidth]{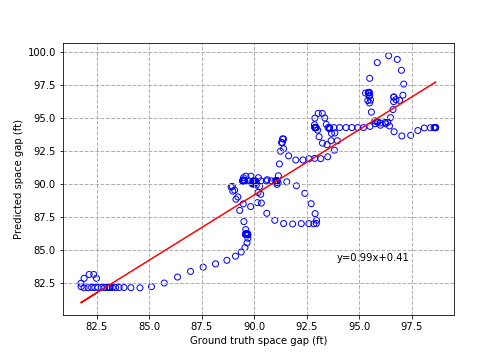}
        \caption{Pipes}
    \end{subfigure}
    \\
    \begin{subfigure}[b]{0.3\textwidth}
        \includegraphics[width=\textwidth]{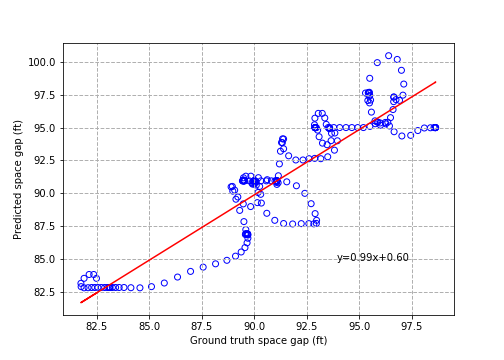}
        \caption{Forbes}
    \end{subfigure}
    \hfill
    \begin{subfigure}[b]{0.3\textwidth}
        \includegraphics[width=\textwidth]{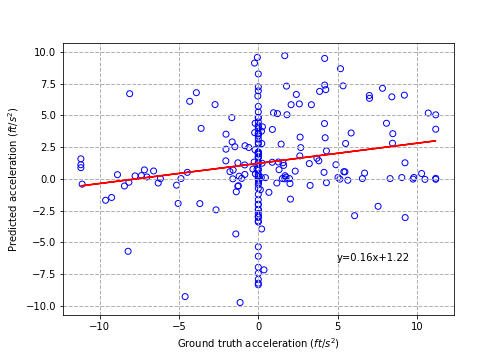}
        \caption{GHR}
    \end{subfigure}
    \hfill
    \begin{subfigure}[b]{0.3\textwidth}
        \includegraphics[width=\textwidth]{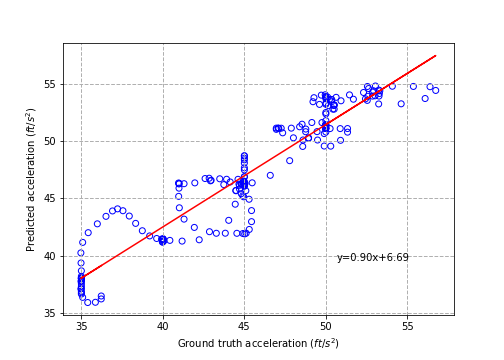}
        \caption{Gipps}
    \end{subfigure}
        \\
    \begin{subfigure}[b]{0.3\textwidth}
        \includegraphics[width=\textwidth]{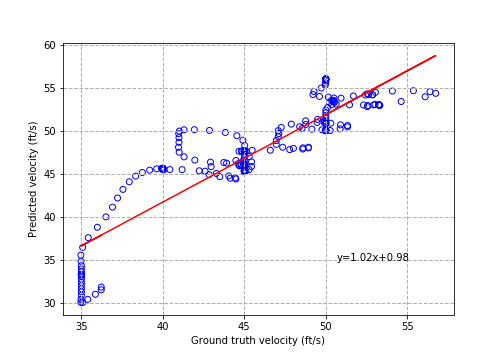}
        \caption{NL}
    \end{subfigure}
    \hfill
    \begin{subfigure}[b]{0.3\textwidth}
        \includegraphics[width=\textwidth]{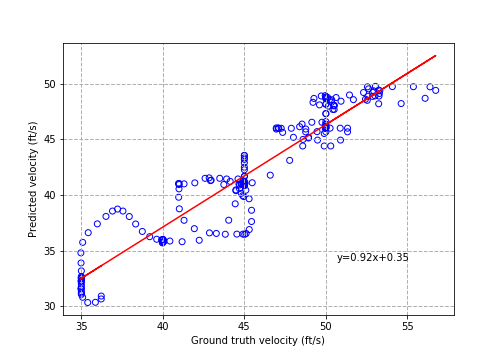}
        \caption{NN}
    \end{subfigure}
    \hfill
    \begin{subfigure}[b]{0.3\textwidth}
        \includegraphics[width=\textwidth]{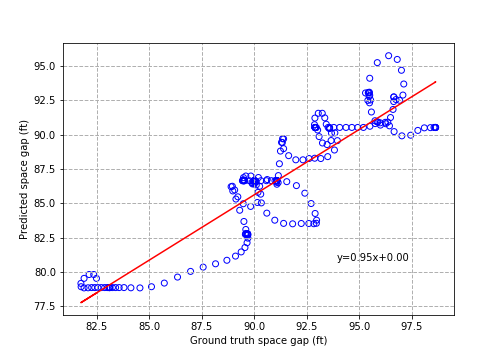}
        \caption{VA}
    \end{subfigure}
    \caption{Comparison between physics model estimation and ground truth without lane changing}
    \label{fig:plot_pk}
       \vspace{-0.2in}
\end{figure}

With the completion of model calibrations, all those physics (car-following) models are implemented to simulate the vehicle trajectories. Figure~\ref{fig:plot_pk} compares the estimation of the predicted quantities listed in Table~\ref{tab:physics} with the ground truth values. If the coefficient of the trend line is close to $1$ and the intercept is close to $0$, the estimation will be considered as accurate. The results show that both vel-DEF and acc-DEF can yield accurate estimations of velocity and accelerations, respectively. However, as the two models didn't consider the car-following behaviors, the obtained good performance can be caused by "over-fitting" and the models cannot be directly applied in the traffic streams that require the modeling the vehicle interactions. Meanwhile, the Gipps model and two Newell models can produce significant estimation errors of vehicle velocity in many cases. In terms of the space gap, Pipes, Forbes, and VA also generate similar levels of performances. For acceleration estimation, the results from GHR is not acceptable. Hence, it can be concluded that the two definition models (Vel-DEF and Acc-DEF) can fit with the input data well, while the car-following behaviors are not captured. All other car-following models tested here have shown obvious estimation errors of velocity, space gap, and acceleration. This is mainly due to the stochastic nature of the vehicle trajectories in the real-world and modeling of such uncertainties goes beyond the capability of deterministic and closed-form physics models.           

\begin{figure}[H]
    \centering
    \begin{subfigure}[b]{0.45\textwidth}
        \includegraphics[width=\textwidth]{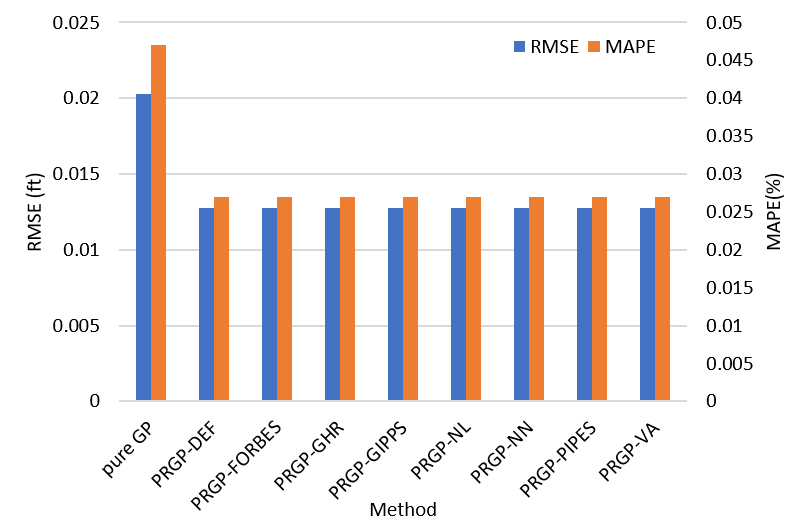}
        \caption{Location x}
    \end{subfigure}
        \hfill
    \begin{subfigure}[b]{0.45\textwidth}
        \includegraphics[width=\textwidth]{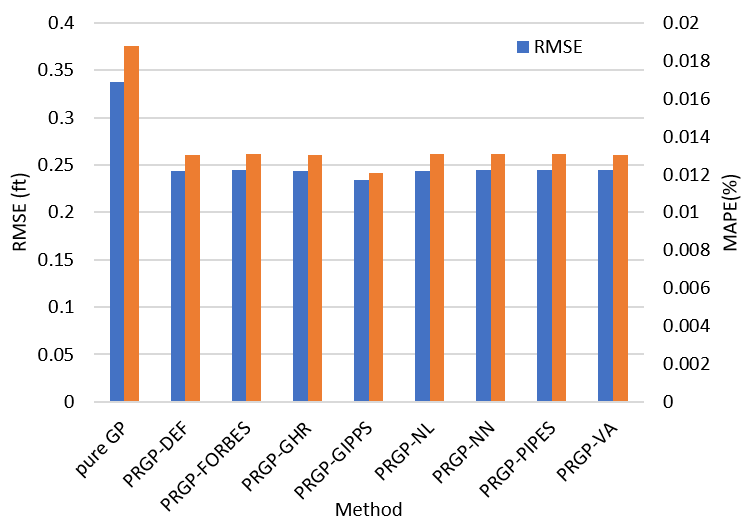}
        \caption{Location y}
    \end{subfigure}
        \\
    \begin{subfigure}[b]{0.45\textwidth}
        \includegraphics[width=\textwidth]{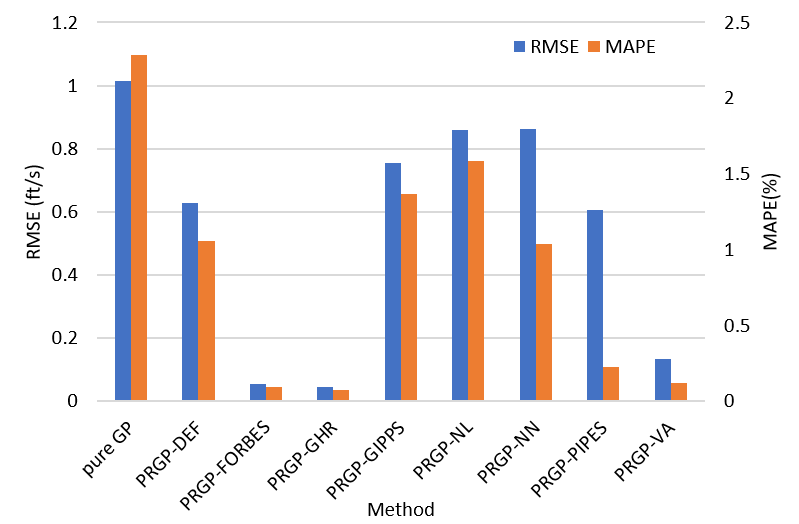}
        \caption{Velocity}
    \end{subfigure}
        \hfill
    \begin{subfigure}[b]{0.45\textwidth}
        \includegraphics[width=\textwidth]{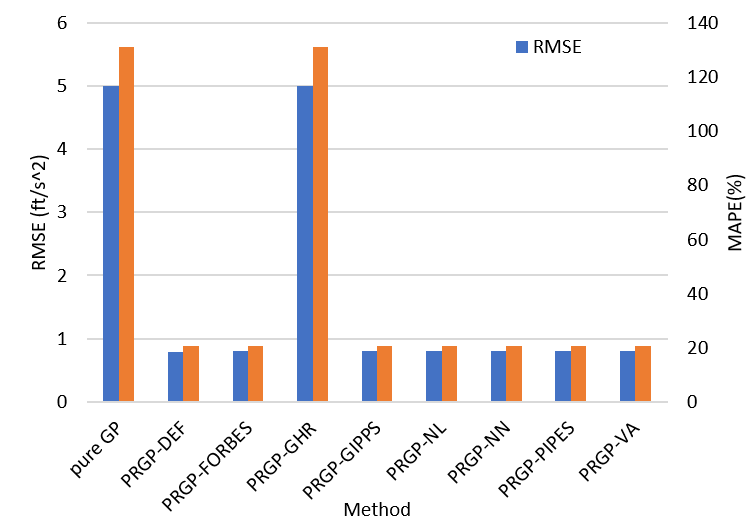}
        \caption{Acceleration}
    \end{subfigure}
    \\
    \begin{subfigure}[b]{0.45\textwidth}
        \includegraphics[width=\textwidth]{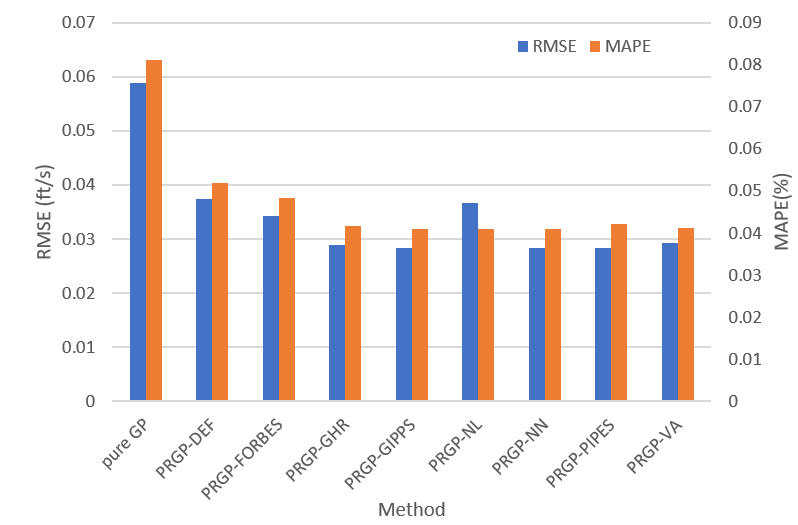}
        \caption{Preceding vehicle velocity}
    \end{subfigure}
    \hfill
    \begin{subfigure}[b]{0.45\textwidth}
        \includegraphics[width=\textwidth]{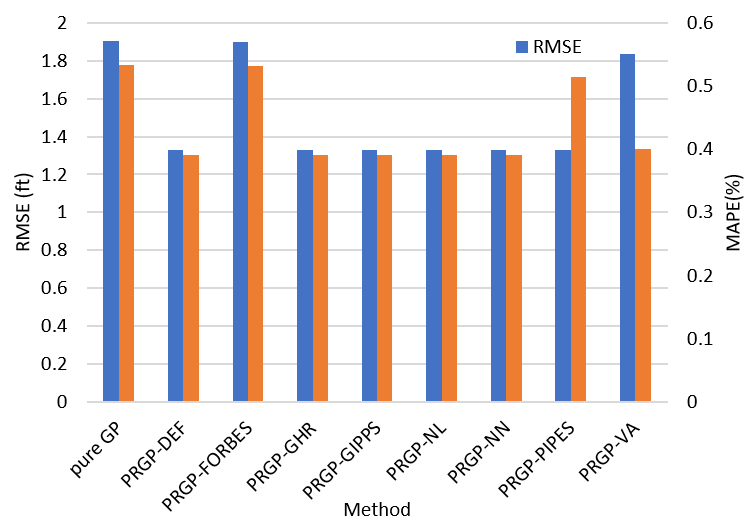}
        \caption{Spatial headway}
    \end{subfigure}
        \\
    \begin{subfigure}[b]{0.45\textwidth}
        \includegraphics[width=\textwidth]{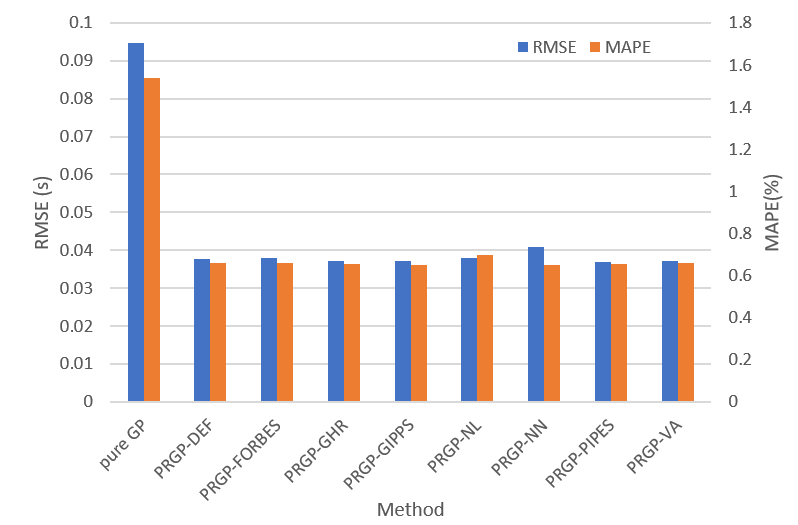}
        \caption{Temporal headway}
    \end{subfigure}

    \caption{Comparison of the performance metrics of Case I}
    \label{fig:plot_metrics_nonoise}
    \vspace{-0.2in}
\end{figure}

Since the tested car-following models have shown their limitations on producing accurate estimation results, this study further adopts the proposed PRGP framework and encodes them into the GP. Also, for the comparison purpose, nine models, including the pure GP and PRGP models as defined in Table~\ref{tab:abbr2}, are trained and tested with the obtained NGSIM trajectory data. Recall that the output of the proposed PRGP models includes each vehicle's lateral and longitudinal coordinates (location x and y), velocity, acceleration, preceding velocity, space gap (spatial headway), and time headway, Figure~\ref{fig:plot_metrics_nonoise} summarizes the estimation RMSE and MAPE of each output variable by each model. Based on the results, all PRGP models yield the same level of performance on estimating the longitude and latitude coordinates of vehicles, but can outperform the pure GP models (see Figure~\ref{fig:plot_metrics_nonoise} (a)-(b)). Note that the estimation errors of 
lateral coordinates - "location x" still exist even though no lane-changing behaviors occurred. This is due to that vehicles do not always stay in the mid of the lane when running. In other words, the "location x" is a continuous variable that ranges from 0 to 1 instead of being a discrete variable. A similar trend can be observed in Figure~\ref{fig:plot_metrics_nonoise} (e) and (g) when estimating the preceding vehicle velocity and temporal headway. In terms of the velocity, PRGP-Forbes, PRGP-GHR, and PRGP-VA produce much smaller RMSE and MAPE of estimations, while the other four PRGP models slightly outperform the pure GP (see Figure~\ref{fig:plot_metrics_nonoise} (c)). For acceleration rate estimation, all PRGP models offers quite accurate results except the PRGP-GHR (see Figure~\ref{fig:plot_metrics_nonoise} (d)). Moreover, Figure~\ref{fig:plot_metrics_nonoise} (f) indicates that PRGP-GHR and PRGP-VA have a similar level of performance on spatial headway estimation as the pure GP while the other PRGP can generate better results. Therefore, it can be concluded that the proposed PRGP models can yield better estimation performances on vehicle trajectories compared with the pure GP. Such results prove the effectiveness of the PRGP when dealing with small training datasets (e.g., only 50 vehicle trajectories were used in this case).

\subsection{Result analysis of Case II}
\begin{figure}[H]
    \centering
    \includegraphics[width=0.7\textwidth]{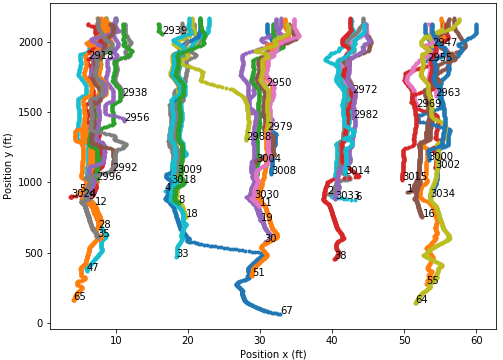}
    \caption{The vehicle trajectory data in the studied case}
    \label{fig:data}
\end{figure}

In this case, this study aims to test the capability of the proposed model to capture the lane-changing behaviors. Fig.~\ref{fig:data} shows the adopted NGSIM vehicle trajectories in the spatial coordinates, where the numbers are the labels of vehicles. Here, the vehicle labels by 2988 and 67 changed the lane and their trajectories will be used for model testing. 

\subsubsection{Model Performance Comparisons}
\begin{figure}[H]
    \centering
    \begin{subfigure}[b]{0.45\textwidth}
        \includegraphics[width=\textwidth]{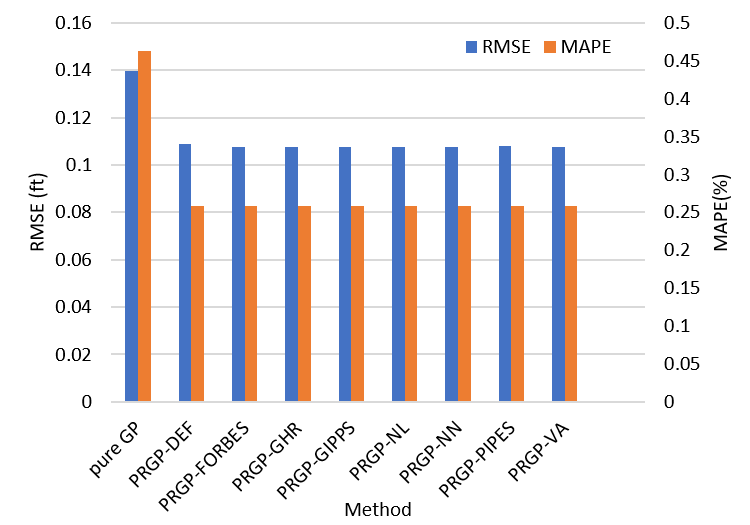}
        \caption{Location x}
    \end{subfigure}
        \hfill
    \begin{subfigure}[b]{0.45\textwidth}
        \includegraphics[width=\textwidth]{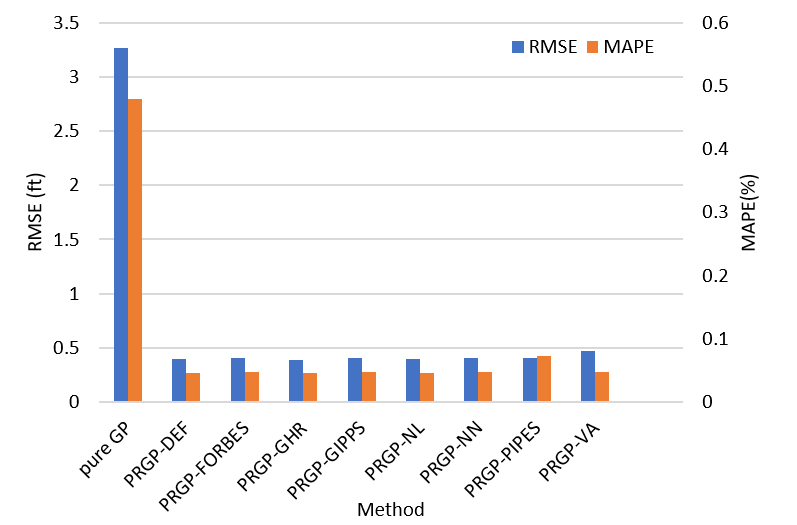}
        \caption{Location y}
    \end{subfigure}
        \\
    \begin{subfigure}[b]{0.45\textwidth}
        \includegraphics[width=\textwidth]{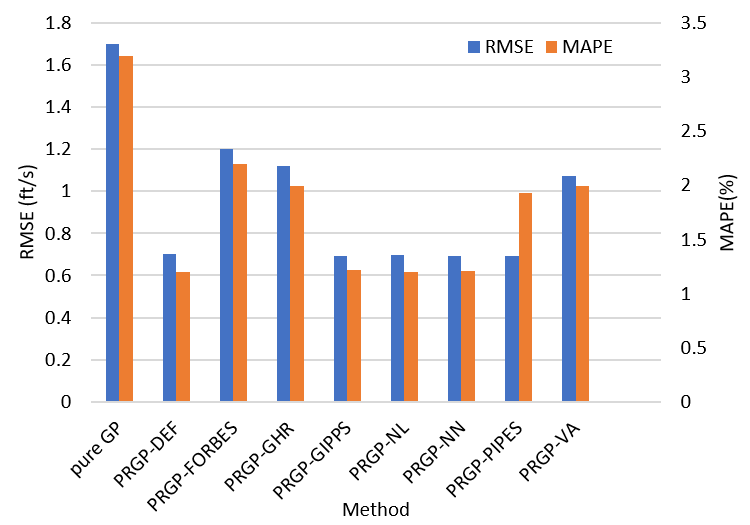}
        \caption{Velocity}
    \end{subfigure}
        \hfill
    \begin{subfigure}[b]{0.45\textwidth}
        \includegraphics[width=\textwidth]{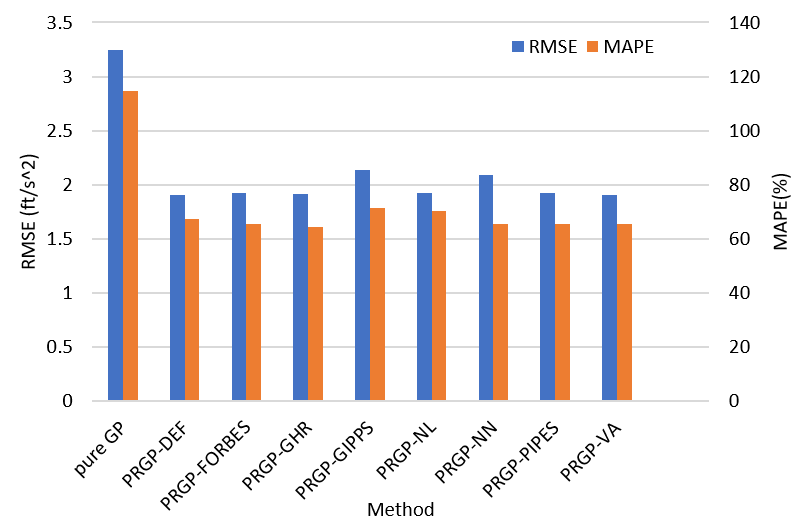}
        \caption{Acceleration}
    \end{subfigure}
    \\
    \begin{subfigure}[b]{0.45\textwidth}
        \includegraphics[width=\textwidth]{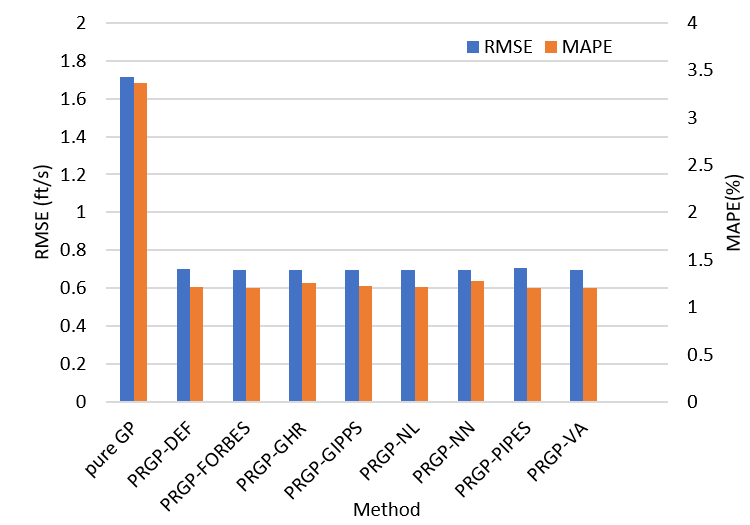}
        \caption{Preceding vehicle velocity}
    \end{subfigure}
    \hfill
    \begin{subfigure}[b]{0.45\textwidth}
        \includegraphics[width=\textwidth]{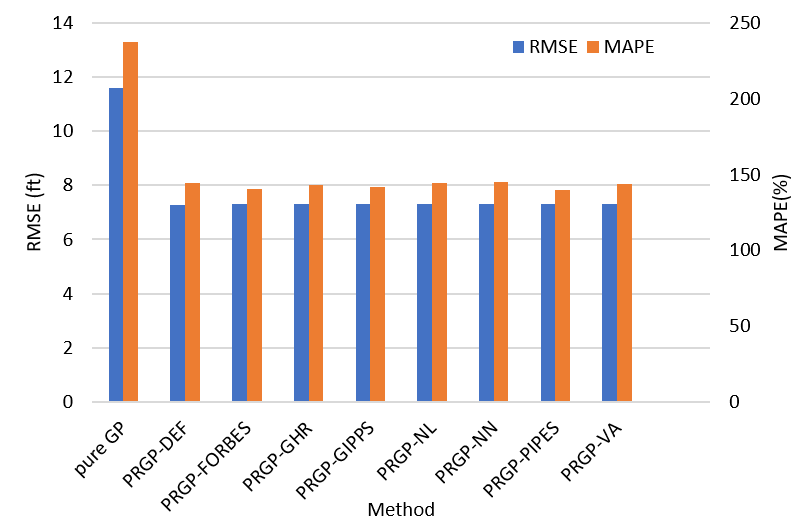}
        \caption{Spatial headway}
    \end{subfigure}
        \\
    \begin{subfigure}[b]{0.45\textwidth}
        \includegraphics[width=\textwidth]{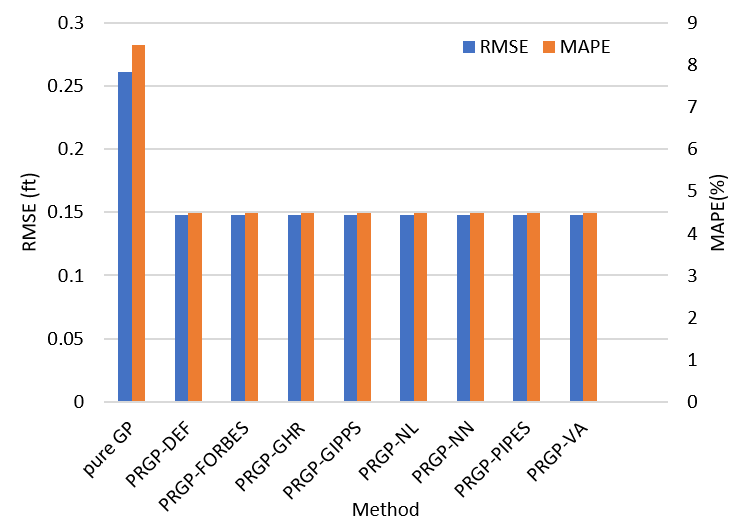}
        \caption{Temporal headway}
    \end{subfigure}
    \caption{Comparison of the performance metrics of Case II}
    \label{fig:plot_metrics_nonoise2}
    \vspace{-0.2in}
\end{figure}

Notably, since only 2 out of 50 vehicles made lane-changing maneuvers in the obtained database, such information will be far away from sufficient for calibrating traditional lane-changing probabilistic models. Hence, in this case, the proposed PRGP models will only be compared with the pure GP model. Taking the output variables, including time-dependent coordinates (i.e., location x and y), velocity, acceleration, preceding vehicle velocity, spatial headway, and temporal heady, as the estimates, Figure~\ref{fig:plot_metrics_nonoise2} presents corresponding RMSE and MAPE of estimations produced by each model. Based on the testing results, it can be observed that the proposed PRGP models yield similar levels of performance on the accuracy, except the estimation of the vehicle's velocity (see Figure~\ref{fig:plot_metrics_nonoise2} (c)). However, all PRGP models can clearly outperform the pure GP in all cases which proves the effectiveness of the PRGP framework. Moreover, the result in Figure~\ref{fig:plot_metrics_nonoise2} (b) indicates that the pure GP yields a relatively large estimation error (i.e., 3.26ft of RMSE) while the ones by the PRGP models are below 0.5ft. The main reason is that pure GP is a data-driven approach and its performance highly relies on the data quantity. In this case, only 2 vehicles made the lane-changing and the data is insufficient to training a reliable machine learning model to study the lane-changing behaviors. By assuming the two vehicles were staying in the same lane, the pure GP consequently leads to a larger estimation error. In contrast, with the integration of the car-following model, the proposed PRGP offers a new solution to deal with the difficulty in model training when the input dataset is small. The integrated car-following knowledge makes the PRGP models more sensitive to the lane-changing records in the data.

\section{Conclusions and Future Research Directions}
Vehicle trajectory data imputation and assimilation are critical and challenging tasks in CAV technologies. The data imputation methods are prone to data noise-induced error. The data assimilation methods are limited by the capability of physics models.
To address these issues, this paper presents a stochastic microscopic traffic model with multivariate Gaussian process (GP) to capture the randomness and measure errors in the real world. A Bayesian inference algorithm is used to estimate the mean and kernel of GPs. A physical knowledge-based regularizer is developed to augment the estimation via a stochastic regularizer. An enhanced latent force model is used to encode physical knowledge into stochastic processes. Based on the posterior regularization inference framework, an efficient stochastic optimization algorithm is developed to maximize the evidence lowerbound of the system likelihood. An empirical study is conducted on the real-world data from the NGSIM dataset. The results show the proposed method outperforms the previous influential methods in estimation precision and has the capability of dealing with car-following and lane-changing behavioral studies concurrently. 

Future research directions include 1) adopting the proposed PRGP models to study the difference of driving behaviors between CAVs and human-driven vehicles; 2) integrating the proposed models into the automated driving systems (ADSs) to improve the safety of CAVs; and 3) implementing the proposed models with a streaming learning framework that can continuously improve the model performance over time when new data are available.

\appendix
\section{Proof of the physical regularized Gaussian Process}\label{app:post_PRGP}
\begin{theorem}
    The physical regularized Gaussian Process is a Gaussian Process.
    \begin{equation}
    p(\mathbf{Y},\omega,\mathbf{g}, \hat{\mathbf{f}},\mathbf{Z}|\mathbf{X})=\mathcal{N(\cdot, \cdot)}
    \label{eq:post_PRGP}
    \end{equation}
    \label{the:1}
\end{theorem}
The parameter hyper-priors $p(\mathbf{Z})$ and $p(\mathbf{\omega})$ are assumed uniformly distributed unless additional domain knowledge of the priors are supplemented.
From Eq.~\ref{eq:post_PRGP}, we marginalize the priors to yield the following equation.
\begin{equation}
    \begin{split}
        p(\mathbf{Y},\omega,\mathbf{g}, \hat{\mathbf{f}},\mathbf{Z}|\mathbf{X})&=p(\mathbf{Z})p(\omega)p(\mathbf{Y},\mathbf{g}, \hat{\mathbf{f}}|\mathbf{X}, \mathbf{Z},\omega)\\
        &\propto p(\mathbf{Y},\mathbf{g}, \hat{\mathbf{f}}|\mathbf{X}, \mathbf{Z},\omega)\\
        &\propto p(\mathbf{Y}|\mathbf{X})p(\hat{\mathbf{f}}|\mathbf{Z})p(\mathbf{g}|\hat{\mathbf{f}})\\
        &=\mathcal{N}(\cdot,\cdot)
    \end{split}
    \label{eq:margin}
\end{equation}
To prove Eq.~\ref{eq:post_PRGP}, we need to prove the conditional probabilities $p(\hat{\mathbf{f}}|\mathbf{Z})$, $p(\mathbf{Y}|\mathbf{X})$, and $p(\mathbf{g}|\hat{\mathbf{f}})$ are Gaussian processes.
From the assumptions, we have the following equations.
\begin{equation}
    p(\mathbf{Y}|\mathbf{X})=\mathcal{N}(\mathbf{Y}|\mathbf{0}, \mathbf{K}+\tau^{-1}\mathbf{I})
    \label{eq:ass1}
\end{equation}
\begin{equation}
    p(\mathbf{Y}|\mathbf{f})=\mathcal{N}(\mathbf{f},\tau^{-1}\mathbf{I})
    \label{eq:ass2}
\end{equation}
To prove $p(\mathbf{g}|\mathbf{f})$ is Gaussian process, we have the following theorem. 
\begin{theorem}
    Given $\Psi$ is a differential operator and $f(\mathbf{x})$ is a GP, $\Psi f(\mathbf{x})$ is another GP.
    \begin{equation}
        \mathbf{f}(\mathbf{x}) = \mathcal{N}(\cdot, \cdot) \iff \Psi \mathbf{f}(\mathbf{x}) = \Psi [\mathcal{N}(\cdot, \cdot)]
        \label{eq:dgp}
    \end{equation}
    \label{the:2}
\end{theorem}
Given $\Psi$ is a linear operator, the variation of Theorem~\ref{eq:twogp} is justified by \citep{alvarez2009latent, alvarez2013linear}.
It is shown that the existence of a unique solution often requires to impose boundary conditions on $\mathbf{f}$, which is expressed by the functional $\mathcal{B}(\mathbf{f}) = \mathbf{c}$ and $\mathbf{c}$ is a constant vector.
The reasoning is based on that applying a linear differential operator on one GP results in another GP \citep{graepel2003solving}.
Given the nonlinear function $\mathbf{f}$ and linear differential operator $\Psi$, \cite{calderhead2009accelerating} found the differential equation was not needed to be solved explicitly.
Given $\Psi$ is a nonlinear operator, the convolution of $\mathbf{f}$ is mostly insolvable.
In these cases, the resultant Gaussian distribution was yield by differentiating the probability distribution of the left-side function as shown in Eq.~\ref{eq:twogp} .
\begin{theorem}
    Given $\Psi$ is a differential operator, if one side of Eq.~\ref{eq:nonlinearlfm} is one GP, the other side is another GP.
    \begin{equation}
        \Psi \mathbf{f}(\mathbf{x}) = \mathcal{N}(\cdot, \cdot) \iff \mathbf{g(x)} = \mathcal{N}(\cdot, \cdot)
        \label{eq:twogp}
    \end{equation}
    \label{the:3}
\end{theorem}
Note that the regularization is fulfilled via a valid generative model component rather than the process differentiation, and hence can be applied to any linear or nonlinear differential operators.
In view of the fact that the resultant covariance and cross-covariance are not obvious via analytical derivatives, the expressive kernels can be learned from data empirically.

Thus, the proposed PRGP is a Gaussian process.\qed

\section{Proof of the algorithm correctness}\label{app:alg}
The posterior regularization is based on optimizing the parameters to maximize the likelihood or the evidence lowerbound (ELBO) of the likelihood.
The objective includes the model likelihood on data and a penalty term that encodes the constraints over the posterior of the variables.
Via the penalty term, we can incorporate our domain knowledge or constrains outright to the posteriors, rather than through the priors and a complex, intermediate computing procedure.
The sum of the evidence lowerbound of likelihoods across the GP and the physical knowledge GP, $\mathcal{L}$, and the proposed inference algorithm is derived as follows.
For convenient effective efficient model inference, we marginalize out all variables in the joint probability to avoid estimating extra approximate posteriors.
Then we derive a convenient evidence lower bound to enable the linear transformation.
Using the auto-differentiation libraries, we develop an efficient stochastic optimization algorithm based on the posterior regularization inference framework.
The generative component is bound to the original GP to obtain a new principled Bayesian model.
The algorithm is based on the maximal likelihood method.
The joint probability is given by Eq.~\ref{eq:yogfzx}.
\begin{equation}
    \label{eq:yogfzx}
    p(\mathbf{Y},\omega,\mathbf{g}, \hat{\mathbf{f}},\mathbf{Z}|\mathbf{X})=p(\mathbf{Y}|\mathbf{X})p(\omega,\mathbf{g},\hat{\mathbf{f}},\mathbf{Z}|\mathbf{X},\mathbf{Y})
\end{equation}
We first marginalize out all the latent variables in the generative component to avoid approximating their posterior in Eq.~\ref{eq:oyx}.
\begin{equation}
    \label{eq:oyx}
    \begin{split}
        p(\omega|\mathbf{X},\mathbf{Y})&=\iiint [p(\omega,\mathbf{g},\hat{\mathbf{f}},\mathbf{Z}|\mathbf{X},\mathbf{Y})\textrm{d}\mathbf{Z}\textrm{d}\mathbf{g}\text{d}\hat{\mathbf{f}}]\\
        &=\iint [p(\mathbf{Z})p(\hat{\mathbf{f}}|\mathbf{Z},\mathbf{X},\mathbf{Y})p(\omega|\Psi\hat{\mathbf{f}},\hat{\mathbf{K}})\textrm{d}\mathbf{Z}\textrm{d}\hat{\mathbf{f}}]\\
        &=\iint [p(\mathbf{Z})p(\hat{\mathbf{f}}|\mathbf{Z},\mathbf{X},\mathbf{Y})\mathcal{N}(\omega|\Psi \hat{\mathbf{f}},\hat{\mathbf{K}})\textrm{d}\mathbf{Z}\textrm{d}\hat{\mathbf{f}}]\\
        &=\mathbb{E}_{p(\mathbf{Z})}\mathbb{E}_{p(\hat{\mathbf{f}}|\mathbf{Z},\mathbf{X},\mathbf{Y})}\mathcal{N}(\Psi \mathbf{\hat{f}}|\mathbf{0},\hat{\mathbf{K}})
    \end{split}
\end{equation}
The prefixed parameter $\gamma \geq 0$ is used to control the strength of regularization effect.
\begin{equation}
    \label{}
    p(\mathbf{Y},\omega|\mathbf{X})=p(\mathbf{Y}|\mathbf{X})p(\omega|\mathbf{X},\mathbf{Y})^\gamma
\end{equation}
The objective is to maximize the log-likelihood in Eq.~\ref{eq:lllh}.
\begin{equation}
    \label{eq:lllh}
    \begin{split}
        \log [p(\mathbf{Y},\mathbf{\omega}|\mathbf{X})]=&\log [p(\mathbf{Y}|\mathbf{X})] + \gamma\log [p(\omega|\mathbf{X},\mathbf{Y})]\\
        =&\log [(\mathcal{N}(\mathbf{Y}|\mathbf{0},\mathbf{\hat{K}}+\tau^{-1} \mathbf{I}))]\\
        &+\gamma \log [\mathbb{E}_{p(\mathbf{Z})} \mathbb{E}_{p(\hat{\mathbf{f}}|\mathbf{Z},\mathbf{X},\mathbf{Y})} [\mathcal{N}(\Psi \hat{\mathbf{f}}|\mathbf{0},\hat{\mathbf{K}})]]
    \end{split}
\end{equation}

However, the log-likelihood is intractable due to the expectation inside the logarithm term.
To address this problem, the Jensen's inequality is used to obtain an evidence lower bound $\mathcal{L}$ in Eq.~\ref{eq:lllhlb}.
\begin{equation}
    \label{eq:lllhlb}
    \begin{split}
        \log [p(\mathbf{Y},\mathbf{\omega}|\mathbf{X})]\geq \mathcal{L}= & \log[\mathcal{N}(\mathbf{Y}|\mathbf{\omega},\hat{\mathbf{K}}+\tau^{-1}\mathbf{I})]\\
        &+\gamma \mathbb{E}_{p(z)} \mathbb{E}_{p(\hat{\mathbf{f}}|\mathbf{Z},\mathbf{X},\mathbf{Y})} [\log [\mathcal{N}(\Psi \hat{\mathbf{f}}|\omega,\hat{\mathbf{K}})]]\\
    \end{split}
\end{equation}

The existence of the general evidence lowerbound (ELBO) of a posterior distribution is proved by analyzing a decomposition of the Kullback-Leibler (KL) divergence by \cite{bishop2006pattern}.
Thus, we can obtain the ELBO of the log-likelihood in Eq.~\ref{eq:lllhlb}. 
However, the ELBO is still intractable due to the non-analytical expectation term $\log [\mathcal{N}(\Psi \hat{\mathbf{f}}|\omega,\hat{\mathbf{K}})]$. 

While the proposed inference algorithm is developed for a hybrid model rather than pure GP \citep{ganchev2010posterior}, the evidence lower bound optimized by Alg.~\ref{alg:1} is a typical posterior regularization objective that estimates a pure GP model and meanwhile penalizes the posterior of the target function to encourage consistency with the differential equations.
Jointly maximizing the term $\mathbb{E}_{p(\mathbf{z})} \mathbb{E}_{p(\hat{\mathbf{f}}|\mathbf{Z},\mathbf{X},\mathbf{Y})} [\log [\mathcal{N}(\Psi \hat{\mathbf{f}}|\omega,\hat{\mathbf{K}})]]$ in the lowerbound of the log-likelihood $\mathcal{L}$ encourages all the possible latent force functions that are obtained from the target function $f(\cdot)$ via the differential operator $\Psi$ should be considered as being sampled from the same shadow GP.
This can be viewed as a soft constraint over the posterior of the target function in the original GP model.
Therefore, while being developed for the inference of a hybrid model, the algorithm is equivalent to estimating the original GP model with some soft constraints on its posterior distribution.
Thus, physical knowledge regularizes the learning of the target function in the original GP.
\qed
\bibliography{mybibfile}

\end{document}